\PassOptionsToPackage{table}{xcolor}
\documentclass{ieeeaccess}

\graphicspath{{Images/}} 
\usepackage[numbers, sort, compress]{natbib}
\usepackage{amsmath,amsfonts}
\usepackage{bm}
\newcommand{\citetitleyear}[1]{\citeauthor{#1} (\citeyear{#1})}

\usepackage{url,lineno,microtype,subcaption}
\usepackage[onehalfspacing]{setspace}
\usepackage[english]{babel}

\newcommand{\tocite}[1]{\textcolor{red}{[MISSING CITATION]}}

\usepackage{mathtools}
\usepackage{amssymb}
\usepackage[inkscapeformat=png]{svg}
\usepackage{pythonhighlight}
  \NewSpotColorSpace{PANTONE}
  \AddSpotColor{PANTONE} {PANTONE3015C} {PANTONE\SpotSpace 3015\SpotSpace C} {1 0.3 0 0.2}
  \SetPageColorSpace{PANTONE}%

\usepackage{graphicx}
\usepackage[export]{adjustbox}
\usepackage{colortbl}
\newcommand{\customrowcolor}{\rownum=0\relax\ifnum\rownum=0\relax\else\ifodd\rownum\relax\cellcolor{gray!30}\fi\fi}

\usepackage{tikz,tikz-3dplot}
\usepackage{pgfplots}
\usepgfplotslibrary{fillbetween}
\usetikzlibrary{calc}

\usepackage{microtype}
\usepackage{graphicx}
\usepackage[colorlinks=True, citecolor=red, linkcolor=.]{hyperref}
\usepackage[utf8]{inputenc} 
\usepackage[T1]{fontenc}    
\usepackage{url}            
\usepackage{booktabs}       
\usepackage{amsfonts}       
\usepackage{color}
\usepackage{mathtools}
\usepackage{amsmath,amssymb}
\usepackage{bm}
\usepackage{adjustbox}

\usepackage{algorithm}
\usepackage{algorithmic}
\usepackage{subcaption}
\usepackage[shortlabels]{enumitem}
\usepackage[capitalise]{cleveref}

\newcommand{\ie}{i.e., }
\newcommand{\eg}{e.g., }

\newcommand{\Skip}[1]{}

\usepackage{array}
\newcommand{\PreserveBackslash}[1]{\let\temp=\\#1\let\\=\temp}
\newcolumntype{C}[1]{>{\PreserveBackslash\centering}p{#1}}
\newcolumntype{R}[1]{>{\PreserveBackslash\raggedleft}p{#1}}
\newcolumntype{L}[1]{>{\PreserveBackslash\raggedright}p{#1}}

\definecolor{red}{RGB}{230, 120, 120}
\definecolor{blue}{rgb}{0.4, 0.6, 0.8}
\definecolor{yellow}{rgb}{1.0, 0.88, 0.21}

\usepackage{amsmath}

\usepackage{pifont, xcolor}
\newcommand*\colourcheck[1]{%
  \expandafter\newcommand\csname #1check\endcsname{\textcolor{#1}{\ding{52}}}%
}
\definecolor{ForestGreen}{RGB}{34, 139, 34}

\colourcheck{blue}
\colourcheck{green}
\colourcheck{ForestGreen}

\usepackage{makecell} 
\usepackage{datatool}

\usepackage{xspace}

\makeatletter
\DeclareRobustCommand\onedot{\futurelet\@let@token\@onedot}
\def\@onedot{\ifx\@let@token.\else.\null\fi\xspace}

\def\eg{\emph{e.g}\onedot} 
\def\ie{\emph{i.e}\onedot}

\makeatother

\usepackage{tablefootnote}

\newcommand{\bracketExist}[1]{%
  \ifx\empty#1\empty
  \else
    \textbf{(#1)}
  \fi
}

\usepackage{duckuments}

\usepackage{subcaption}
\DeclareMathOperator*{\minimize}{min}
\DeclareMathOperator*{\find}{find}
\DeclareMathOperator*{\subjto}{subj. to}
\DeclarePairedDelimiterX{\norm}[1]{\lVert}{\rVert}{#1}

\usepackage{multirow}

\def\BibTeX{{\rm B\kern-.05em{\sc i\kern-.025em b}\kern-.08em
    T\kern-.1667em\lower.7ex\hbox{E}\kern-.125emX}}
    
\begin{document}

\history{Date of publication xxxx 00, 0000, date of current version xxxx 00, 0000.}
\doi{10.1109/ACCESS}

\title{A Review of Differentiable Simulators}
\author{\uppercase{Rhys Newbury}\authorrefmark{1,2}, \uppercase{Jack Collins}\authorrefmark{3}, \uppercase{Kerry He}\authorrefmark{1},
\uppercase{Jiahe Pan}\authorrefmark{4}, \uppercase{Ingmar Posner}\authorrefmark{3}, \uppercase{David Howard}\authorrefmark{5}, and \uppercase{Akansel Cosgun}\authorrefmark{6}
\address[1]{Monash University, Australia}
\address[2]{Australian National University, Australia}
\address[3]{University of Oxford, United Kingdom}
\address[4]{University of Melbourne, Australia}
\address[5]{CSIRO, Brisbane, QLD 4069, Australia}
\address[6]{Deakin University, Australia}
\tfootnote{This work was supported by a UKRI/EPSRC Programme Grant [EP/V000748/1].}}

\markboth
{Newbury \headeretal: A Review of Differentiable Simulators}
{Newbury \headeretal: A Review of Differentiable Simulators}


\corresp{Corresponding author: Rhys Newbury (e-mail: rhys.newbury@monash.edu).}

\begin{abstract}

Differentiable simulators continue to push the state of the art across a range of domains including computational physics, robotics, and machine learning. Their main value is the ability to compute gradients of physical processes, which allows differentiable simulators to be readily integrated into commonly employed gradient-based optimization schemes. To achieve this, a number of design decisions need to be considered representing trade-offs in versatility, computational speed, and accuracy of the gradients obtained. This paper presents an in-depth review of the evolving landscape of differentiable physics simulators.
We introduce the foundations and core components of differentiable simulators alongside common design choices. This is followed by a practical guide and overview of open-source differentiable simulators that have been used across past research. Finally, we review and contextualize prominent applications of differentiable simulation. By offering a comprehensive review of the current state-of-the-art in differentiable simulation, this work aims to serve as a resource for researchers and practitioners looking to understand and integrate differentiable physics within their research. We conclude by highlighting current limitations as well as providing insights into future directions for the field.


\end{abstract}

\begin{keywords}
Differentiable Simulator, Review, Differentiable Physics, Soft Body Simulation, System Identification, Trajectory Optimization, Morphology Optimization, Policy Optimization, Robotics
\end{keywords}

\titlepgskip=-21pt

\maketitle


\section{Definition}

\subsection{Gradients}

\subsection{Dynamical Model}

\subsection{Integrator}

\begin{itemize}
    \item \textbf{Euler Integrator}: The simplest integrator, Euler's method, updates an object's position and velocity by taking a small time step and applying the current velocity to the position and the current acceleration to the velocity. While easy to implement, Euler's method can lead to stability and accuracy issues, particularly with more complex simulations.
    
    \item \textbf{Semi-Implicit Euler Integrator}: This integrator improves upon the standard Euler method by first updating the velocity using the current acceleration and then updating the position using the updated velocity. It provides slightly better stability but might still exhibit issues with energy conservation and accuracy.
    
    \item \textbf{Runge-Kutta Integrators}: Runge-Kutta methods, particularly the fourth-order Runge-Kutta (RK4), offer higher accuracy and stability compared to basic Euler methods. These integrators use multiple stages to update positions and velocities and are widely used in many physics engines for accurate simulations.
    
    \item \textbf{Implicit Integrators}: Implicit integrators are designed to handle stiff differential equations and constraints more effectively. They use iterative techniques to solve for the new positions and velocities in each time step. Common examples include the Backward Euler and Implicit Midpoint methods.

    \item \textbf{Multi-Step Integrators}: Multi-step methods like the Adams-Bashforth and Adams-Moulton methods use information from multiple previous time steps to update positions and velocities. These can offer higher accuracy but might require more memory and computation.
    
\end{itemize}

\subsection{Coordinate Sytem} 

Minimal and maximal coordinate systems are two different ways of representing the configuration of a multi-body system in simulation.

\begin{itemize}
    \item Minimal coordinates use a set of generalized coordinates to represent the position and orientation of all the bodies in the system. These coordinates are typically chosen to be independent of each other, which makes the equations of motion easier to solve. However, this can lead to problems with numerical stability, especially when the system is close to a configuration singularity.
    \item Maximal coordinates use the full six degrees of freedom for each body in the system. This means that the equations of motion are more complex, but they are also more robust to numerical errors. Additionally, maximal coordinates make it easier to handle closed kinematic loops and nonholonomic constraints.
\end{itemize}

In general, minimal coordinates are more efficient to compute, but maximal coordinates are more robust. 

\subsection{Type} - Engine + Application / Application

\subsection{Application Type} -

\subsection{Open Source} -

\subsection{Documentation} -

\subsection{Actively Maintained} -

\subsection{Language} -

\section{Introduction}
\label{sec:intro}

Physics simulators are extensively utilized within the sciences, and are a key enabling technology for a range of industrial, design, engineering, and robotics applications~\cite{collins-2021-a-applications}. These simulators are grounded in mathematical models of physical laws, coupled to pertinent constraints, \eg joint and velocity limits for robotics applications. This allows for a principled approach to numerically predict forward in time given the current system state.

Simulators have increasingly leveraged rapidly evolving hardware resources (\ie multi-threaded CPUs and GPUs, high performance compute clusters) to execute parallel simulations far faster than real-time, allowing for large amounts of data to be collected simultaneously. This has made simulation a key tool for machine learning applications, particularly in scenarios where data is not readily available, such as robotics~\cite{Mittal2023Orbit, openai2019solving}. 

\begin{figure*}[htbp!]%
\centering
\includegraphics[width=1.0\linewidth]{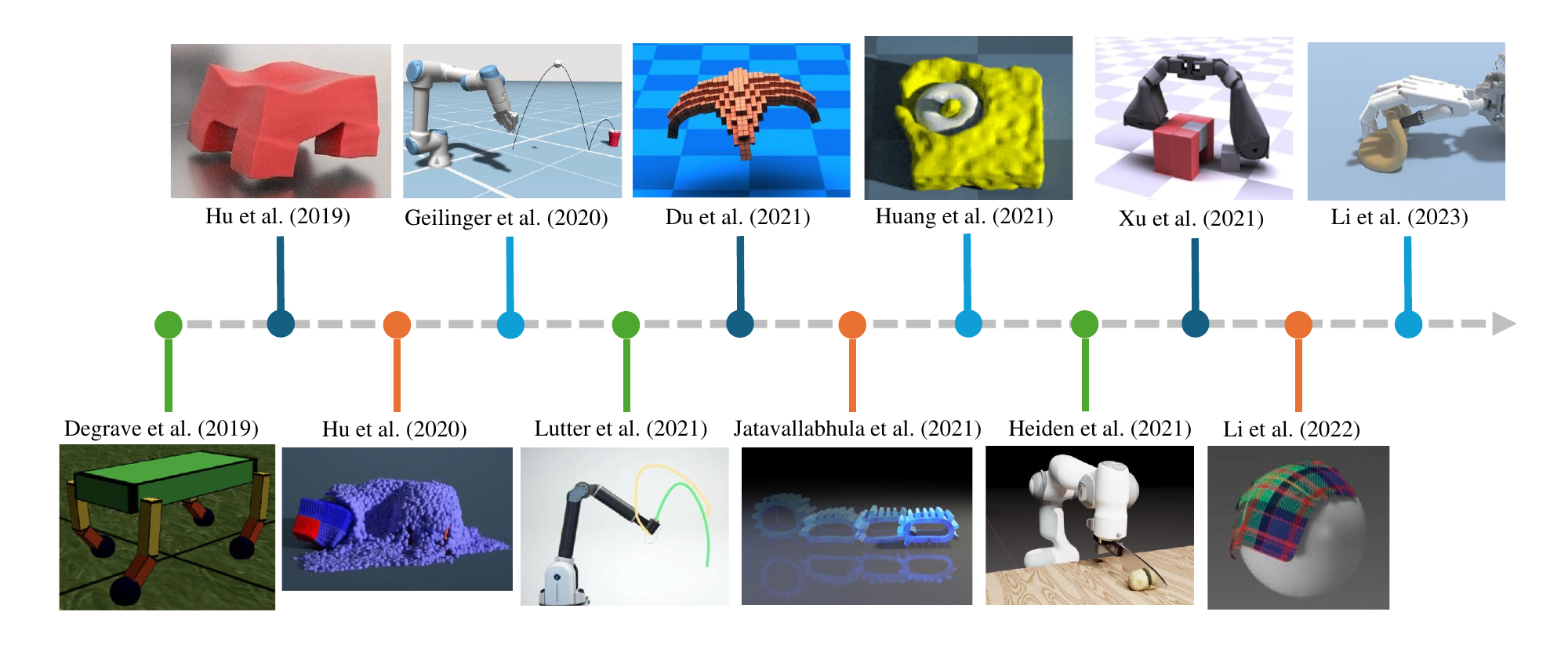}
\caption{A visualization of the breadth of published research progressing the field of differentiable simulation. Areas and applications of differentiable simulators cover topics such as soft and rigid-body simulation; 
system identification; trajectory optimization; morphology optimization; and many others covered in-detail in this review \cite{degrave-2019-a-robotics,hu-2020-difftaichi-simulation,geilinger-2020-add-contact,du-2021-diffpd-dynamics,jatavallabhula-2021-gradsim-control,huang-2021-plasticinelab-physics,heiden-2021-disect-cutting,Xu-2021-An-Design,li-2023-dexdeform-physics,li-2022-diffcloth-contact} and \cite{hu-2019-chainqueen-robotics,lutter-2021-differentiable-learning} {\small(\textsuperscript{\textcopyright} 2024 IEEE)}}
\label{fig:collage}
\vspace{-0.15cm}
\end{figure*}

Despite providing computationally-tractable data generation, traditional physics simulators critically do not provide access to the gradient information that machine learning heavily relies on to drive the learning process. This limitation has spurred the development of a new class of \textit{differentiable} simulators (see Fig.~\ref{fig:collage} for a visualization of the breadth of published differentiable simulator research). Differentiable simulators are end-to-end differentiable, \ie capable of calculating gradients throughout the entire duration of a simulation for a given loss function with respect to desired parameters (\eg surface friction with respect to a mean squared error loss comparing the simulated trajectory and a ground-truth trajectory). Thus, they provide a direct route for integration within gradient-based optimization frameworks and deep learning pipelines. Since the initial exploration of a general-purpose differentiable simulator by \citet{degrave-2019-a-robotics}, the field has grown rapidly, to the point where there are several well-supported and maintained differentiable physics simulators.

Recognizing the emergence of this field of research, this review aims to provide a timely and comprehensive overview of differentiable simulators -- what they are, how they work, and how have they been used in research thus far. Additionally, we curate an up-to-date list of currently supported and well-maintained differentiable physics simulators, serving as a reference for practitioners when selecting suitable options for their specific applications. To ensure comprehensive coverage, this review encompasses all relevant literature published before 2024 that proposes or utilizes differentiable physics simulators, with some necessary caveats which we will outline in the relevant sections. In summary, the contributions of this paper are, (i) a thorough review of the fundamental concepts and methodologies of differentiable physics simulators, (ii) a curated list of well-maintained differentiable simulators to aid practitioners in selecting appropriate tools, (iii) applications of differentiable simulators demonstrating how these simulators have been used in the past, and (iii) a discussion on future directions of the field.

The review comprises the following sections: Foundations of Differentiable Simulators (\cref{sec:differentiable_physics}), which discusses the core components of a differentiable physics simulator and what makes them differentiable; Differentiable Simulators (Section~\ref{sec:differentiable_simulators}), covering a subset of open-source differentiable simulators utilized in past research, and therefore this selection offers a good starting point for exploring the practical side of differentiable simulators; Applications (Section~\ref{sec:applications}) including system identification, trajectory optimization, policy optimization and morphology optimization; and, Discussion (Section~\ref{sec:discussion}) with a future outlook of the field of differentiable simulators.

%


\section{Foundations of Differentiable Simulators}
\label{sec:differentiable_physics}

\begin{figure}[htbp!]
    \centering
    \includegraphics[width=\linewidth]{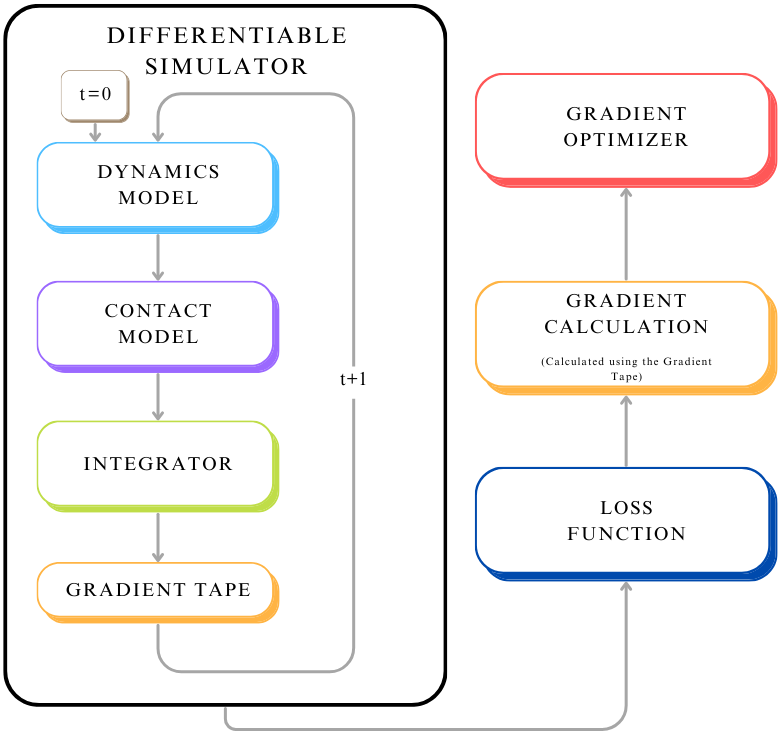}
    \caption{An overview of how the different components of a differentiable simulator interact. Each of these components (except for the loss function and gradient optimizer) are explored in detail in \cref{sec:differentiable_physics}.}
    \label{fig:block_diagram}
\end{figure}

A differentiable simulator is comprised of several core components. 
\begin{enumerate}
\item Gradient Calculation -- this is the component of differentiable simulators which distinguishes them from standard physics simulators, and is responsible for computing gradients with respect to the simulator's parameters.
\item Dynamics Model -- the underlying physics governing the simulated system. For the purposes of this review, we restrict our attention to simulators whose physics are governed by a predefined set of equations based on physical kinematic or dynamic constraints.
\item  Contact Model -- most differentiable simulators implement a contact model to simulate interactions between objects during collision events. These contact models introduce a unique challenge to calculating gradients due to the inherently discontinuous nature of contacts.
\item Integrator -- an integrator numerically solves the equations of motion over discrete time steps. Although the derivation and computation of gradients through explicit integration schemes is fairly straightforward, implicit integration schemes, which usually require solving a nonlinear system of equations to obtain the state at the next time step, introduce unique challenges.
\end{enumerate}
These components collectively form the underlying substrate of a differentiable physics simulator. Fig.~\ref{fig:block_diagram} presents a simplified visualization of how the different components interact to produce gradients of a physics simulation that can be used for optimization.

The scope of this review is differentiable physics simulators with the exclusion of neural networks trained on simulated data, \eg~\cite{battaglia-2016-interaction-physics, chang-2017-a-dynamics, mrowca-2018-fllexible-prediction, schenck-2018-spnets-networks, sanchez-2020-learning-networks, li-2019-propagation-observation, de-2020-combining-prediction, ummenhofer-2020-lagrangian-convolutions, wandel-2021-learning-generalize}, as although they provide a method of deriving gradients that are physics informed, they are not constrained to conform to the underlying physics model. Additionally, we exclude traditional simulators employing finite differencing for derivative calculations, such as \cite{todorov-2012-mujoco-control}, as purpose-built differentiable simulators offer derivatives with speed and accuracy not achievable by traditional simulators. Excluding such avenues for calculating physics-informed gradients ensures a more focused examination of the techniques used when developing the main identified components of differentiable simulators, without diluting the primary points of interest. However, we acknowledge that this reduced scope of our review will exclude some seminal work in the field of differentiable simulation.

\begin{filecontents*}{Tables/table_data.csv}
Year,Paper,SimName,GradientMethod,DynamicalModel,SoftBody,InternalForces,Integrator,ContactModel
2019,degrave-2019-a-robotics,,Auto Diff,Newton,-,-,Semi-Implicit,Complementarity
2018,de-2018-end-control,,Auto Diff,Newton,-,-,Explicit,LCP
2019,hu-2019-chainqueen-robotics,,\Gape[0pt][2pt]{\makecell[c]{Analytical\\Symbolic}},Continuum Mechanics,\ForestGreencheck,MLS-MPM,Explicit,MLS-MPM
2019,liang-2019-differentiable-problems,,Implicit Diff,Newton,\ForestGreencheck,FEM,Implicit,Position Based
2020,geilinger-2020-add-contact,,Adjoint Method,Newton,\ForestGreencheck,Neo-Hokean,Implicit,NCP
2021,heiden-2021-neural-networks,,Auto Diff,Newton,-,-,Semi-Implicit,NCP + Compliant
2021,freeman-2021-brax-body,Brax,Newton,XPBD,-,-,Semi-Implicit,Position Based
2020,qiao-2020-scalable-control,,Implicit Diff,Newton,-,-,Implicit,Position Based
2021,jatavallabhula-2021-gradsim-control,GradSim,Auto Diff,Newton,\ForestGreencheck,-,Semi-Implicit,Compliant Model
2021,huang-2021-plasticinelab-physics,,Auto Diff,Continuum Mechanics,\ForestGreencheck,MLS-MPM,Explicit,MLS-MPM
2021,werling-2021-fast-contact,Nimble,Symbolic,Lagrangian,-,-,Explicit,LCP
2019,heiden-2019-interactive-simulation,TDS,Auto Diff,Newton,-,-,Semi-Implicit,Compliant Model
2021,du-2021-underwater-simulation,,Auto Diff,Projective Dynamics,\ForestGreencheck,Corotated Linear,Implicit,Complementarity
2023,le-2023-differentiable-objects,,Auto Diff,Newton,-,-,Implicit,Compliant Model
2021,ding-2021-dynamic-language,,Auto Diff,Newton,-,-,Implicit,Impulse Based
2021,lutter-2021-differentiable-learning,,Auto Diff,Newton,-,-,Not Specified,None
2021,xu-2021-accelerated-simulation,,Auto Diff,Newton,-,-,Semi-Implicit,Compliant Model
2021,qiao-2021-efficient-bodies,,Adjoint Method,Newton,-,-,Explicit,LCP
2020,song-2020-learning-simulations,,Analytical,Newton,-,-,Explicit,None
2021,zhong-2021-extending-models,,Implicit Diff,Lagrangian,-,-,Explicit,Convex Optimization
2021,mora-2021-pods-simulation,,Adjoint Method,Newton,-,-,Implicit,None
2021,scibior-2021-imagining-simulation,,Auto Diff,Newton,-,-,Not Specified,None
2022,turpin-2022-grasp-hands,,Auto Diff,Newton,-,-,Semi-Implicit,Compliant Model
2023,turpin-2023-fast-simulation,,Auto Diff,Newton,-,-,Semi-Implicit,Position Based
2021,wang-2021-sim2sim-robots,,Auto Diff,Newton,-,-,Semi-Implicit,Impulse Based
2022,gartner-2022-differentiable-reconstruction,,Auto Diff,Newton,-,-,Semi-Implicit,LCP
2021,le-2021-differentiable-identification,,Implicit Diff,Lagrangian,-,-,Implicit,Complementarity
2023,liu-2023-robotic-dynamics,,Auto Diff,XPDB,\ForestGreencheck,Potential Energy,Explicit,Position Based
2022,howelll-2022-dojo-robotics,Dojo,Implicit Diff,Lagrangian,,,Implicit,Complementarity
2022,wang-2022-recurrent-data,,Auto Diff,Newton,-,-,\Gape[0pt][2pt]{\makecell[c]{Implicit\\Semi-Implicit}},Impulse Based
2023,spielberg-2023-advanced-chainqueen,,Auto Diff,Continuum Mechanics,\ForestGreencheck,MLS-MPM,Explicit,MLS-MPM
2022,nava-2022-fast-models,,Auto Diff,\Gape[0pt][2pt]{\makecell[c]{Fluid Sim\\Continuum Mechanics}},\ForestGreencheck,Linear Corotated,Implicit,None
2022,gong-2022-fine-fabrics,,Implicit Diff,Lagrangian,\ForestGreencheck,Elastic Rod,Implicit,Compliant Model
2022,granados-2022-model-physics,,Auto Diff,Newton,-,-,Explicit,None
2022,wang-2022-differentiable-musculotendons,,Auto Diff,Newton,-,-,Implicit,None
2023,wiedemann-2023-training-gradient,,Auto Diff,Newton,-,-,Explicit,None
2019,heiden-2019-real2sim-physics,,Auto Diff,Newton,-,-,Semi-Implicit,None
2022,petrik-2022-learning-physics,,Auto Diff,Newton,-,-,Explicit,Impulse Based
2023,chen-2023-daxbench-physics,daX,Auto Diff,Continuum Mechanics,\ForestGreencheck,MLS-MPM,Explicit,MLS-MPM
2022,zhao-2022-graph-robots,,Analytical,\Gape[0pt][2pt]{\makecell[c]{Newton\\Fluid Simulation}},-,-,Not Specified,None
2022,zhao-2022-automatic-grammar,,Auto Diff,\Gape[0pt][2pt]{\makecell[c]{Newton\\Fluid Simulation}},-,-,Explicit,None
2021,du-2021-diffpd-dynamics,,Adjoint Method,Projective Dynamics,\ForestGreencheck,Potential Energy,Implicit,Complementarity
2020,holl-2020-phiflow-simulations,,Adjoint Method,Fluid Simulation,-,,Explicit,None
2023,wang-2023-softzoo-environments,,Auto Diff,Continuum Mechanics,\ForestGreencheck,MLS-MPM,Semi-Implicit,MLS-MPM
2023,DiffXPBD-stuyck-2023-dynamics,,Analytical,XPBD,\ForestGreencheck,,Compliant Model,Implicit
2020,holl-2020-phiflow-simulations,PhiFlow,Auto Diff,Fluid Simulation,-,-,None,Explicit
2023,bezgin-2023-JAX-flows,JAX-Flows,Auto Diff,Fluid Simulation,,,None,Explicit
2023,xian-2023-fluidlab-manipulation,FluidLab,Analytical,{\makecell[c]{Continuum Mechanics\\Fluid Simulation}},,,Explicit,MLS-MPM
\end{filecontents*}
\DTLloaddb{mydata}{Tables/table_data.csv} 
\DTLsort{Year,Paper}{mydata} 
\begin{table*}[h!]
\caption{A table of all the differentiable physics simulators found throughout our review. This does not include every paper within the scope of the review, only those papers which present a new engine for differentiable simulation, not application-based papers which use existing engines. We highlight important components of differentiable simulators as columns within the table. For more details on each column, see \cref{subsec:gradient} for Gradient Methods, \cref{subsec:dynamics} for Dynamical Models, \cref{subsec:contact} for Contact Models, and \cref{subsec:integrator} for Integrators.}
\label{tab:background}
\rowcolors{2}{gray!30}{white}
\centering
\resizebox{0.90\textwidth}{!}{%
\begin{tabular}{|l|c|c|c|c|c|}
\hline
Paper & Gradient Method & Dynamical Model & \makecell[c]{Soft\\Body} & Contact Model & Integrator \DTLforeach*{mydata}{\Citation=Paper,\SimName=SimName,\GradientMethod=GradientMethod,\DynamicalModel=DynamicalModel,\SoftBody=SoftBody,\InternalForces=InternalForces,\ContactModel=ContactModel,\Integrator=Integrator} {%
\DTLiffirstrow{\\\hline}{\\}\citetitleyear{\Citation}   \bracketExist{\SimName} & \GradientMethod & \DynamicalModel & \SoftBody & \ContactModel & \Integrator}\\\hline\end{tabular}}
\vspace{5mm}
\end{table*}

Throughout this section, $q,\dot{q},\ddot{q}\in\mathbb{R}^n$ represent the $n$-dimensional generalized coordinates of the system, their time derivatives, and second time derivates (accelerations), respectively. Many sources do not explicitly state parts of their methodology, so where possible the code was examined to try and ascertain the correct information. Papers where the required information could not be sourced are not included in the following subsections. An overview of all the differentiable physics engines covered in this review coupled with the high-level design choices made for each is included in \cref{tab:background}.


\subsection{Gradient Calculation}\label{subsec:gradient}



There are many methods to calculate gradients. In the following, we will review automatic differentiation in \cref{subsubsec:ad}, symbolic differentiation in \cref{subsubsec:symbolic}, and analytic derivations of gradients in \cref{subsubsec:analytical}.

The majority of simulation frameworks reviewed in this paper use automatic differentiation as the backbone for their gradient computations. This is because automatic differentiation avoids the need to manually derive gradients by hand, which can be increasingly cumbersome the more complex the simulator is. In comparison, symbolic differentiation, despite also automating the evaluation of the gradient of a given function, needs to build symbolic expressions of the gradient which grow exponentially with the size of the function being differentiated~\cite{baydin2018automatic}, and is therefore rarely used over automatic differentiation in differentiable simulators. Both of these automated gradient calculation tools incur an overhead which can be disadvantageous in performance critical applications. Similarly, it can be difficult to apply these methods on dynamics which are discontinuous or given as, \eg, a solution of a feasibility problem rather than an explicit function. Both of these issues arise when modelling contacts (see Section~\ref{subsec:contact}). In these scenarios, it can be advantageous to analytically derive expressions for the gradients instead, or to use analytical techniques (\eg, from linear algebra) to simplify the computation of the gradient. Overall, we are aware of only one work which does not use automatic differentiation~\cite{song-2020-learning-simulations}, instead calculating all gradients analytically.

\subsubsection{Automatic Differentiation} \label{subsubsec:ad}

Automatic differentiation (AD) relies on the fact that even complex functions are composed of elementary operations (\eg, addition, subtraction, etc.) and functions (\eg, $\sin$, $\cos$, $\exp$, etc.), which allows for repeatedly applying the chain rule to these expressions in a programmatic way to compute the value of a derivative. There are two main types of AD: forward mode and reverse mode. Forward mode AD traverses through the chain rule starting from the function input, and is more efficient when there are more outputs than inputs. In comparison, reverse mode AD starts from the function output, and is more efficient in cases where there are more inputs than outputs~\cite{baydin2018automatic}. As differentiable simulators are typically used to find the derivative of a (scalar valued) objective function with respect to many simulation parameters, reverse mode AD is typically used for these applications. An alternative approach to AD is source code transformation, which is used by ~\citet{hu-2020-difftaichi-simulation}. This uses a just-in-time (JIT) transpiler over Python code to produce a function which will calculate gradients. 



Many differentiable simulators~\cite{de-2018-end-control,heiden-2021-neural-networks,jatavallabhula-2021-gradsim-control,huang-2021-plasticinelab-physics,heiden-2019-interactive-simulation,du-2021-underwater-simulation,le-2023-differentiable-objects,ding-2021-dynamic-language,lutter-2021-differentiable-learning,xu-2021-accelerated-simulation,scibior-2021-imagining-simulation,turpin-2022-grasp-hands,wang-2021-sim2sim-robots,gartner-2022-differentiable-reconstruction,liu-2023-robotic-dynamics,wang-2022-recurrent-data,spielberg-2023-advanced-chainqueen,nava-2022-fast-models,granados-2022-model-physics,wang-2022-differentiable-musculotendons,wiedemann-2023-training-gradient,heiden-2019-real2sim-physics, degrave-2019-a-robotics, holl-2020-phiflow-simulations, bezgin-2023-JAX-flows} use a pipeline based \emph{solely} on AD, commonly using either PyTorch Autograd~\cite{paszke-2017-automatic-pytorch}, JAX~\cite{jax2018github} or DiffTaichi~\cite{hu-2020-difftaichi-simulation}. 



\subsubsection{Symbolic Differentiation}\label{subsubsec:symbolic}

Like AD, symbolic differentiation computes the derivative of a function by applying the chain rule repeatedly. However, whereas AD computes a numerical value for the derivative, symbolic differentiation produces a symbolic expression. However, for applications using differentiable simulators, usually we are only interested in the gradient evaluated at a single point to perform gradient based optimization, and therefore the full symbolic gradient is not required. Additionally, as previously discussed, the main drawback of symbolic differentiation is that the size of the symbolic gradients grow rapidly with the complexity of the function. Therefore, symbolic differentiation is typically only used when the dynamics are simple, or for a few select components of the simulator. This is done in~\cite{werling-2021-fast-contact, geilinger-2020-add-contact, hu-2019-chainqueen-robotics}.

\subsubsection{Analytical Gradients} \label{subsubsec:analytical}




It can be challenging to apply AD and symbolic differentiation to discontinuous processes or phenomena. In these situations, it is advantageous to \emph{manually} derive explicit expressions for derivatives of functions arising from differentiable simulators. Also, while AD and symbolic differentiation require the computation of gradients online, analytical gradients provide an explicit expression for the gradient a priori, allowing gradients to be computed more quickly and accurately. Another situation where it can be difficult to apply AD and symbolic differentiation is when expressions are given as a feasibility problem, which often arises in contact modeling (see~\cref{subsec:contact}). Such components often also require analytic derivations of gradients.

The gradients of an entire simulator can be derived manually. For example, \cite{song-2020-learning-simulations} derive analytical expressions for an environment with a sliding object. However, this can be very cumbersome for more complex scenarios, which involve many bodies with complex contact dynamics. In comparison, \citep{zhao-2022-graph-robots, hu-2019-chainqueen-robotics} only manually derive gradients for performance critical components of their system.

When manually deriving the explicit expressions for gradients, authors often follow standard techniques such as implicit differentiation or the adjoint method. These methods will be discussed in the remainder of this subsection.


\textbf{Implicit Differentiation: } Implicit differentiation is a technique used to obtain a derivative $dy/dx$ from an expression $f(y, x) = 0$ containing the two variables, even when one variable is not explicitly represented as a function of the other, \ie, as $y(x)$. This technique can be particularly useful for contact models which are expressed as feasibility or optimization problems, which can be difficult to obtain gradients for using AD or symbolic differentiation.

\citet{liang-2019-differentiable-problems} and \citet{gong-2022-fine-fabrics} use implicit differentiation to derive the gradients for a Newtonian and Lagrangian dynamic model, respectively. \citet{howelll-2022-dojo-robotics} and~\citet{le-2023-differentiable-objects} use the implicit function theorem on a set of optimality conditions to compute the gradients of a contact model given as the solution of a feasibility problem.



In OptNet~\citep{amos-2017-optnet-networks}, it was shown how implicit differentiation could be used to obtain gradients through an expression given by a Quadratic Program (QP), which are optimization problems of the form
\begin{subequations}\label{eqn:qp}
    \begin{align}
        \minimize_{z} \quad & \frac{1}{2} z^T Qz + q^Tz\\
        \subjto \quad & Az = b\\
        & Gz \leq h. \label{eqn:qp-c}
    \end{align}
\end{subequations}
Subsequent research has extended this formulation to address diverse constraints. \citet{liang-2019-differentiable-problems} refine the implicit differentiation technique of a QP to minimize the resulting size of the linear system. This is achieved using QR decomposition, a well-known matrix factorization method. Expanding on this, \citet{qiao-2020-scalable-control} further extends the methodology to handle nonlinear constraints by incorporating information from the Jacobian. Additionally, \citet{le-2021-differentiable-identification} extends these methods to address quadratically constrained quadratic programs (QCQPs), \ie, problems of the form \cref{eqn:qp} but with the linear inequality constraint \cref{eqn:qp-c} replaced with $p$ quadratic constraint $z^\top G_i z + g_i^\top z \leq h_i$ for $i=1,\ldots,p$. Furthermore, the gradients for a linear complementarity problem (LCP) formulation (see \cref{subsub:complementarity} for more detail) can also be derived using similar techniques~\cite{de-2018-end-control}.

\textbf{Adjoint Method: } The adjoint method leverages an adjoint vector obtained by solving a linear system to make the process of obtaining gradients less computationally expensive. This adjoint vector encapsulates information about how changes in the system's output (\eg, the final state) relate to changes in the parameters. Using this adjoint vector in the gradient calculation allows efficient calculation of parameters that influence the system behavior without directly computing how each parameter affects the entire trajectory. 

More concretely, consider the derivative of a loss function $\mathcal{L}$ with respect to some parameters $\theta$
\begin{equation}\label{eqn:adjoint-a}
    \frac{d\mathcal{L}}{d\theta} = \frac{\partial\mathcal{L}}{\partial\theta} + \frac{\partial\mathcal{L}}{\partial q}^\top\frac{dq}{d\theta}.
\end{equation}
The main challenge is in computing $\frac{dq}{d\theta}$, which can be obtained by implicitly differentiating through the system dynamics $f(q, \theta)=0$. However this is expensive as it involves solving a large number of linear equations. The adjoint method avoids this by first computing the adjoint vector $z=(\frac{\partial f}{\partial q})^{-\top}\frac{\partial \mathcal{L}}{\partial q}$ by solving a single linear system, then computes the desired gradient as
\begin{equation}
    \frac{d\mathcal{L}}{d\theta} = \frac{\partial\mathcal{L}}{\partial\theta} - z^\top\frac{\partial f}{\partial\theta}.
\end{equation}
This method is adopted by~\cite{qiao-2021-efficient-bodies,geilinger-2020-add-contact,du-2021-diffpd-dynamics,li-2022-diffcloth-contact,Xu-2021-An-Design,DiffXPBD-stuyck-2023-dynamics} for computations of gradients of the loss function. In~\cite{mora-2021-pods-simulation}, an extension of the adjoint method by~\cite{zimmermann2019optimal} was used to obtain the Hessian of the loss function, which can be used to accelerate the convergence of optimization algorithms.

\subsection{Dynamics Model}\label{subsec:dynamics}

A dynamics model is a mathematical model that describes the behavior of a system over time. It can be used to predict how the system will respond given changes to its inputs. There are distinct approaches when modeling dynamic systems, with three prominent categories: rigid body dynamics, soft body dynamics and fluid dynamics. Rigid body and soft body dynamics are strongly linked. However, a key distinction is that soft body dynamics incorporates internal forces within the dynamics model. This inclusion enables a more faithful representation of the complex interactions and deformation characteristics of soft materials. In rigid body dynamics, Newtonian or Lagrangian methods are typically used for modeling the system's dynamics. However, these methods can be extended to handle soft bodies with the inclusion of internal forces. There are other dynamics models which are purpose-built for modeling soft body objects, such as continuum mechanics~\cite{hu-2019-chainqueen-robotics}, Projective Dynamics~\cite{bouaziz-2023-projective-simulation}, or Compliant Position-Based Dynamics (XPBD)~\cite{macklin-2016-xpbd-dynamics}. Fluid dynamics, distinct from rigid and soft body dynamics, involves the study of fluid flow patterns and interactions within fluidic environments.

\subsubsection{Rigid Body Dynamics}

The \textbf{Newtonian} dynamics model is based on Newton's laws of motion, which expresses the acceleration of a body as a function of the forces acting on it and its mass, \ie,
\begin{equation}\label{eqn:newton}
F = m\ddot{q},
\end{equation}
where $F$ is the force acting on the body, $m$ is the mass of the body, and $\ddot{q}$ is the acceleration of the body. However, this neglects the rotational motion of the body and only considers the translational motion. Newton-Euler takes into account the rotational motion of the body, \ie,
\begin{equation}
\label{eqn:newton-euler}
    M(q)\ddot{q} + C(q, \dot{q}) = \tau,
\end{equation}
where $\tau$ is a vector of generalized forces, $M$ is the mass matrix, and $C$ is the bias force matrix which accounts for other forces acting on the system, including centrifugal forces, Coriolis forces and gravity. 

Several approaches use Newton or Newton-Euler dynamics to model their rigid-body system~\cite{degrave-2019-a-robotics,de-2018-end-control,freeman-2021-brax-body,qiao-2020-scalable-control,heiden-2019-interactive-simulation,le-2023-differentiable-objects,lutter-2021-differentiable-learning,xu-2021-accelerated-simulation,song-2020-learning-simulations,mora-2021-pods-simulation,turpin-2023-fast-simulation,wang-2021-sim2sim-robots,gartner-2022-differentiable-reconstruction,wang-2022-recurrent-data,wang-2022-differentiable-musculotendons,wiedemann-2023-training-gradient,heiden-2019-real2sim-physics,zhao-2022-graph-robots,zhao-2022-automatic-grammar,geilinger-2020-add-contact,liang-2019-differentiable-problems}. \citet{granados-2022-model-physics} apply Newton's method but only consider the steady-state response arising from the model.

Many works incorporate kinematic constraints into Newtonian dynamics for their specific problem domain. For example, \citet{turpin-2022-grasp-hands} use a slight extension to Newtonian dynamics to account for the kinematics of a robotic hand. \citet{heiden-2021-neural-networks, qiao-2021-efficient-bodies} uses the Articulated Body Algorithm~\cite{featherstone-2014-rigid-algorithms} which is based on Newton-Euler recursive dynamics. \citet{scibior-2021-imagining-simulation} adopt a simple differentiable model, a kinematic bicycle. 

Some works frame Newtonian dynamics as an impulse-based model rather than a force-based model. \citet{ding-2021-dynamic-language, petrik-2022-learning-physics} make use of an impulse-based model, using conservation of momentum to calculate the change in velocities when there are collisions of objects. \citet{jatavallabhula-2021-gradsim-control} updates momentum in terms of $F = \frac{dp}{dt}$, where $p$ is the linear momentum of an object. An integration step is then performed to calculate the new momentum and the updated linear velocity can be computed as $v = p / m$. \\

The \textbf{Lagrangian} dynamics formulation is based on the principle of least action, which states that the motion of a system will minimize the action, which is a function of the system's configuration and its time derivatives.

The Lagrangian for a simple particle is given by
\begin{equation}
\mathcal{L} = T - V,
\end{equation}
where $T$ is the kinetic energy of the particle, and $V$ is the potential energy of the particle. The Euler-Lagrange equation, which describes the equations of motion for a system in terms of its Lagrangian, is given by
\begin{equation}
    \frac{d}{dt}\biggl(\frac{\partial \mathcal{L}}{\partial \dot{q}}\biggr) - \frac{\partial \mathcal{L}}{\partial 
 q} = 0.
\end{equation}
Lagrangian dynamics are adopted by \cite{werling-2021-fast-contact,le-2021-differentiable-identification, gong-2022-fine-fabrics, chen-2022-real-simulation, lelidec-2022-augmenting-smoothing}. In comparison, \citet{howelll-2022-dojo-robotics} derive their physics by using Hamilton's Principle of Least-Action. By directly performing integration on these equations, the simulator is able to automatically conserve momentum and energy in the system.
It is important to note that Newton, Lagrangian, and Hamiltonian dynamics are all equivalent models~\cite{morin-2008-introduction-solution} but provide different ways to formulate the equations.\\

\textbf{Extending Rigid Body to Soft Body: } 
To extend rigid body dynamics to soft bodies, additional considerations need to be included for internal forces within the soft body. These internal forces address how the soft body resists changes in shape, volume, and overall configuration. Internal forces have been used in both Newton dynamical model~\cite{geilinger-2020-add-contact, jatavallabhula-2021-gradsim-control, liang-2019-differentiable-problems} and Lagrangian models~\cite{gong-2022-fine-fabrics}. A Neo-Hookean material model is used in~\cite{geilinger-2020-add-contact, jatavallabhula-2021-gradsim-control} which describes the energy density as a function of the deformation gradient. An FEM method is adopted by~\cite{liang-2019-differentiable-problems}, where for each triangle face in the mesh, the deformation gradient is calculated as a variable of the strain. \citet{gong-2022-fine-fabrics} model cloth as woven elastic rods, and derive the equations of force from ~\cite{jawed-2018-a-rods}.

\subsubsection{Soft Body} \label{subsubsec:soft}

\textbf{Continuum Mechanics} deals with the mechanical behavior of materials modeled as continuous substances, rather than as collections of discrete particles. It provides a framework for analyzing the motion and deformation of solids and fluids. The governing equations of continuum mechanics are typically expressed in terms of the conservation of mass and momentum~\citep{hu-2018-a-coupling}. Momentum conservation, also known as Newton's Second Law for a Continuous Material, states that the change in momentum of a material element is equal to the sum of the divergence of the stress tensor and the gravitational forces acting on the element, \ie,
\begin{equation}
\rho \frac{Dv}{Dt} = \nabla \cdot \sigma + \rho g,
\end{equation}
where $\rho$ represents the material density, $v$ is the velocity of the particle, $\sigma$ is the Cauchy stress tensor and $\frac{D\phi}{Dt} \equiv \frac{\partial\phi}{\partial t} + v \dot \nabla (q)$ is the material derivative of any function $\phi(x, t)$. Mass conservation states that the rate of change of density plus the divergence of the mass flux (density multiplied by velocity) is zero, \ie,
\begin{equation}
\frac{D\rho}{Dt} + \rho \nabla \cdot v = 0.
\end{equation}
Continuum mechanics is adopted by MLS-MPM models~\cite{hu-2019-chainqueen-robotics,huang-2021-plasticinelab-physics,spielberg-2023-advanced-chainqueen,chen-2023-daxbench-physics,wang-2023-softzoo-environments}. MLS-MPM~\cite{hu-2018-a-coupling} computes the deformation gradient by reusing quantities computed from the Affine Particle-In-Cell Method~\cite{jiang-2015-the-method}. Simpler models for internal forces have also been adapted by~\cite{chen-2023-daxbench-physics} who use a mass-spring system to model a cloth. The MLS-MPM model is extended by \cite{huang-2021-plasticinelab-physics} to include a von Mises yield criterion for modeling plasticity~\cite{gao-2017-an-materials}. According to the von Mises yield criterion, a plasticine particle deforms permanently if the stress exceeds a certain limit, and a correction is needed to account for the material's changed initial state. \cite{heiden-2021-disect-cutting} use a different model for continuum mechanics, using the Finite Element Method (FEM) to compute elastic forces based on a Neo-Hookean constitutive model. This model considers material properties such as the Young’s modulus and Poisson’s ratio, while aiming to preserve volume during large deformations.

\textbf{Projective Dynamics} is an implicit integration method (see \cref{subsubsec:implicit}) used in computer graphics which simulates the dynamics of deformable objects \cite{bouaziz-2023-projective-simulation}. Projective Dynamics models object deformation as a projection onto its rest state, then minimizes an energy function that accounts for internal and external forces to obtain the desired dynamics.

Project Dynamics is extended by DiffPD~\citep{du-2021-diffpd-dynamics} to be differentiable and then integrated into a differentiable simulator. In the context of differentiable simulators, DiffPD provides a faster method for performing implicit integration while requiring less computational memory. This model is then adapted into other work such as \cite{du-2021-underwater-simulation, li-2022-diffcloth-contact}.

\citet{li-2022-diffcloth-contact} define the internal forces as $f_{int} = -\nabla E$, where $E$ is the potential energy function, this is based on the original definition proposed by \cite{bouaziz-2023-projective-simulation}. In contrast to this, a co-rotated linear material model~\cite{sifakis-2012-fem-reduction} is used by \cite{du-2021-diffpd-dynamics, du-2021-underwater-simulation}. This model employs a strain measure derived from a polar decomposition and features an energy function that includes terms accounting for both linear and nonlinear characteristics in the material's response to deformation.

\textbf{Compliant Position-Based Dynamics} (XPBD)~\cite{macklin-2016-xpbd-dynamics} is a method for simulating the dynamic behavior of deformable objects. XPBD uses a position-based approach~\cite{muller-2007-position-dynamics}, directly computing particle positions instead of calculating velocities and subsequently updating positions. These positions are solved iteratively to satisfy given internal (\eg, distances between particles on the same body) and external constraints (\eg, rigid contacts, external forces). The internal constraints are an alternative approach to calculating internal forces, by constraining the particles according to physical laws, realistic deformations and interactions can be ensured. For example, \citet{liu-2023-robotic-dynamics} defines a series of internal constraints for a rope (Shear and Stretch, Bend and Twist, Distance) to ensure realism for the simulated rope.

XPBD is adopted by two differentiable simulation engines, Warp~\citep{macklin-2022-warp-graphics} and Brax~\citep{freeman-2021-brax-body} as well as \cite{liu-2023-robotic-dynamics}. However, Brax only uses XPBD for rigid body dynamics, not using any internal constraints. These simulators make use of automatic differentiation libraries to calculate the gradient of the dynamics. Recently, \citet{DiffXPBD-stuyck-2023-dynamics} show how gradients of the XPBD framework can be derived analytically.

\subsubsection{Fluid Simulation}

Some works also aim to bring differentiable capabilities to the domain of fluid simulation. To model hydrodynamics, \cite{zhao-2022-graph-robots, zhao-2022-automatic-grammar, du-2021-underwater-simulation} use aerodynamic force equations for both drag and lift. They discretize geometry to a triangular mesh and compute the lift and drag forces on each surface triangle. To model fluid flow, 
\cite{nava-2022-fast-models, holl-2020-phiflow-simulations} design differentiable simulation engines which aim to solve the incompressible Navier-Stokes equations~\cite{boyer-2012-mathematical-models}. \citet{um-2020-solver-solvers} studies advection-diffusion models which describe the temporal evolution of velocity in a fluid system, influenced by advection, diffusion, and external forces, as well as being subject to additional equality constraints. \citet{bezgin-2023-JAX-flows} aim to design a differentiable framework for computational fluid dynamics (CFD) which allows for the simulation of complex phenomena such as three-dimensional turbulence, compressibility effects, and two-phase flow.

\subsection{Contact model}\label{subsec:contact}

One of the main difficulties in implementing a differentiable physics simulator is how to model contacts, as these tend to exhibit highly nonlinear and discontinuous behavior. These discontinuities arise as contacts exist in a binary state, as the objects can only ever be in contact or not. This can be seen in \cref{fig:contacts} where contact forces are discontinuous, step-like functions, where the gradient is not well-defined. 

In the following, contact forces will be denoted as $\lambda=(\lambda_n, \lambda_t)\in\mathbb{R}\times\mathbb{R}^2$, where $\lambda_n$ and $\lambda_t$ are the normal and tangential forces relative to the contact surface between two objects. Similarly, relative velocities between two objects are represented as $\dot{q}=(\dot{q}_n, \dot{q}_t)\in\mathbb{R}\times\mathbb{R}^2$, where $\dot{q}_n$ and $\dot{q}_t$ are the normal and tangential velocities relative to the contact surface between two objects.

Contact models typically serve two purposes: to resolve interpenetrating objects, and to apply frictional forces. First, the contact model ensures that two solid objects in contact cannot penetrate through each other. This non-penetration requirement can be mathematically represented as
\begin{align}\label{eqn:normal}
    (\lambda_n, \dot{q}_n) \in \{ (\lambda_n, \dot{q}_n)\in\mathbb{R}\times\mathbb{R} : \lambda_n \geq 0,  \dot{q}_n \geq 0,  \lambda_n \dot{q}_n = 0 \},
\end{align}
\ie contact forces should always repel two objects instead of pulling them together, two objects cannot penetrate each other, and objects are either in contact (normal velocities are zero) or are not in contact (normal forces are zero). This last condition is known as a complementarity condition. 

Second, frictional forces should be modeled when two objects are in contact with each other. Frictional forces in physics simulators are most commonly modeled using Coulomb's friction law. This states that the maximum magnitude of the frictional force is proportional to the normal force, \ie
\begin{equation}\label{eqn:friction-cone}
    \lambda \in \mathcal{K}_f = \{ (\lambda_n, \lambda_t) \in \mathbb{R}_+ \times \mathbb{R}^2 : \norm{\lambda_t}_2 \leq \mu \lambda_n \},
\end{equation}
where $\mu\geq0$ is the coefficient of friction between two surfaces. Note that $\mathcal{K}_f$ is a (nonlinear) second-order cone. Additionally, friction must always act in the opposite direction to the tangential velocity, \ie
\begin{equation}\label{eqn:friction-dir}
    \lambda_t = -\mu \lambda_n \frac{\dot{q}_t}{\norm{\dot{q}_t}_2}, \quad \mathrm{if} \quad \norm{\dot{q}_t}_2>0.
\end{equation}
We refer the reader to~\cite{horak-2019-on-simulation,zhong-2022-differentiable-control,lidec-2023-contact-analysis} for further resources on contact models.

\subsubsection{Complementarity problem}
\label{subsub:complementarity}
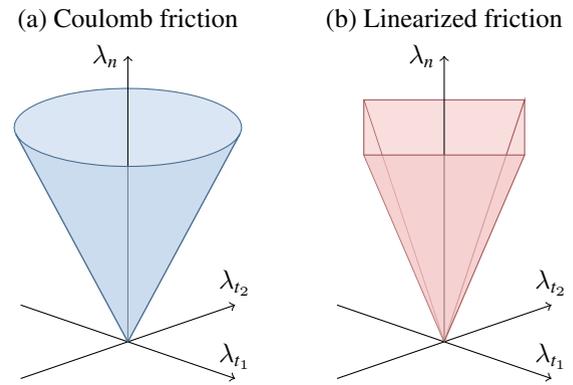
\begin{figure}[]
\centering
\tdplotsetmaincoords{70}{45}
\tdplotsetrotatedcoords{-90}{180}{-90}

\tikzset{surface/.style={draw=blue!70!black, fill=blue!40!white, fill opacity=.6}}

\newcommand{\coneback}[4][]{
    \draw[canvas is xy plane at z=#2, #1] (45-#4:#3) arc (45-#4:225+#4:#3) -- (O) --cycle;
}
\newcommand{\conefront}[4][]{
    \draw[canvas is xy plane at z=#2, #1] (45-#4:#3) arc (45-#4:-135+#4:#3) -- (O) --cycle;
}

\begin{tikzpicture}[tdplot_main_coords]
  \coordinate (O) at (0,0,0);

  \draw[->] (-2,0,0) -- (2,0,0) node[above] {$\lambda_{t_1}$};
  \draw[->] (0,-2,0) -- (0,2,0) node[above] {$\lambda_{t_2}$};
  \coneback[surface]{3}{1.5}{10}
  \draw[->] (O) -- (0,0,4) node[left] {$\lambda_n$};
  \conefront[surface]{3}{1.5}{10}

  \node[] at (0,0,4.5) {(a) Coulomb friction};
\end{tikzpicture}\hspace{0.75cm}\begin{tikzpicture}[tdplot_main_coords]
    \coordinate (O) at (0,0,0);
    \coordinate (A1) at (1.5,0,3);
    \coordinate (A2) at (0,-1.5,3);
    \coordinate (A3) at (-1.5,0,3);
    \coordinate (A4) at (0,1.5,3);
    
    \draw[->] (-2,0,0) -- (2,0,0) node[above] {$\lambda_{t_1}$};
    \draw[->] (0,-2,0) -- (0,2,0) node[above] {$\lambda_{t_2}$};
  
    \draw[draw=red!70!black, fill=red!40!white, fill opacity=.6] (O)--(A3)--(A4);
    \draw[draw=red!70!black, fill=red!40!white, fill opacity=.6] (O)--(A4)--(A1);
    \draw[draw=red!70!black, fill=red!40!white, fill opacity=.6] (O)--(A2)--(A3);
    
    \draw[->] (O) -- (0,0,4) node[left] {$\lambda_n$};
    \draw[draw=red!70!black, fill=red!40!white, fill opacity=.6] (O)--(A1)--(A2);
  
    \node[] at (0,0,4.5) {(b) Linearized friction};
  
\end{tikzpicture}
\caption{Comparison between (a) the second-order frictional cone defining Coulomb's law (see \cref{eqn:friction-cone}), and (b) the square pyramid which linearizes the cone for LCP. Note that the linearized cone biases frictional forces towards the edges of the pyramid.}
\label{fig:cones}
\end{figure}


The most natural way to solve for contact forces is to find a suitable pair of forces and velocities that satisfy the non-penetration conditions and Coulomb's friction law. This involves solving the following feasibility problem
\begin{subequations}
    \begin{align}
        \find \quad &(\lambda, \dot{q}) \in \mathbb{R}^3\times\mathbb{R}^3 \\
        \subjto \quad & \eqref{eqn:normal},\ \eqref{eqn:friction-cone},\ \mathrm{and}\ \eqref{eqn:friction-dir}.
    \end{align}
\end{subequations}
This is referred to as a nonlinear complementarity problem (NCP), where the nonlinearities arise from the frictional constraints. Some works solve this NCP, or variations of this NCP, directly~\cite{geilinger-2020-add-contact,howelll-2022-dojo-robotics,heiden-2021-neural-networks}. Alternatively, some works~\cite{de-2018-end-control,degrave-2019-a-robotics,werling-2021-fast-contact,qiao-2021-efficient-bodies,gartner-2022-differentiable-reconstruction} instead approximate the second-order friction cone (\cref{fig:cones}(a)) as a friction pyramid (\cref{fig:cones}(b)), which results in a linear complementarity problem (LCP). Although easier to solve, a drawback to this linearization is that frictional forces are biased towards the corners of the friction pyramid~\cite{horak-2019-on-simulation,lidec-2023-contact-analysis}.

To obtain gradients from the complementarity-based contact model, works which use optimization methods~\cite{geilinger-2020-add-contact,howelll-2022-dojo-robotics,de-2018-end-control}, such as interior-point methods, can derive the gradients by implicitly differentiating through the Karush-Kuhn-Tucker (KKT) conditions (see \cref{subsubsec:analytical}). The KKT conditions are necessary optimality criteria for constrained optimization problems. Another common way the complementarity problem can be solved is by using the projected Gauss-Seidel (PGS) method~\cite{heiden-2021-neural-networks,degrave-2019-a-robotics,qiao-2021-efficient-bodies}. A simple method to obtain gradients through this method is to automatically differentiate through the PGS algorithm~\cite{heiden-2021-neural-networks,degrave-2019-a-robotics}, although, this can result in a large computational overhead. Conventional implicit differentiation techniques on the optimality conditions can avoid building this computational graph by directly obtaining explicit expressions for the gradient. However, \citet{qiao-2021-efficient-bodies} notes that the PGS method does not guarantee that the complementarity constraints are satisfied when the algorithm is terminated, and therefore implicit differentiation techniques can produce inaccurate gradients. Instead, \citet{qiao-2021-efficient-bodies} proposes a reverse version of the PGS method using the adjoint method (see Section~\ref{subsubsec:analytical}) to obtain gradients, without requiring the computational overhead of automatic differentiation techniques. In~\citet{werling-2021-fast-contact}, which solves the LCP using standard linear programming techniques, gradients are derived analytically without recasting it as an optimization problem. 


Alternatively, in~\cite{du-2021-diffpd-dynamics,li-2022-diffcloth-contact}, the complementarity conditions are directly incorporated into the Projective Dynamics framework (see Section~\ref{subsubsec:soft}). 
The Projective Dynamics solver is augmented with an active-set method which iteratively searches for which constraints are active. Gradients are then obtained using a similar method to the Projective Dynamics framework without contacts (see the last paragraph of \cref{subsubsec:implicit}).




Related to the complementarity problem formulation, \cite{zhong-2021-extending-models,le-2021-differentiable-identification} approximate the complementarity problem as a convex optimization problem (more specifically, as a quadratic program), this approach takes inspiration from the approach in MuJoCo~\cite{todorov-2012-mujoco-control}. Gradients are returned by implicitly differentiating through the optimality conditions of the convex optimization problem.


A related class of contact models to complementarity problems are impulse-based methods, which similarly account for contacts by changing state velocities. Impulse-based methods include~\cite{ding-2021-dynamic-language,petrik-2022-learning-physics,wang-2022-recurrent-data,wang-2022-real2sim2real-engine,wang-2021-sim2sim-robots}, which model non-penetration requirements as an explicit impulse function which are easy to automatically differentiate over. 

For these methods, it was identified in~\cite{hu-2020-difftaichi-simulation} that gradients at collision points can be inaccurate when the model is discretized in time. In particular, a na\"ive time discretization assumes that collisions only occur at discrete time intervals, and this inaccuracy worsens with larger time intervals. To tackle this issue, a continuous time-of-impact (TOI) method is proposed in~\cite{hu-2020-difftaichi-simulation} which computes the precise time at which a contact occurs, even if the contact occurs in between two discrete time intervals. In~\cite{zhong-2022-differentiable-control}, the TOI method was shown to produce more accurate contact gradients for a variety of complementarity- and impulse-based contact models. Other variations of the TOI method are proposed by~\cite{zhong-2023-improving-contacts,wang-2022-recurrent-data} aiming to produce more accurate gradients.

\subsubsection{Compliant models}

\begin{figure*}[]
\centering
\pgfmathdeclarefunction{TrueFriction}{1}{%
  \pgfmathparse{%
    (and(   1,    #1<0)*(1)            +%
    (and(#1>= 0,    1  )*(-1)%
    }%
}

\pgfmathdeclarefunction{LinFriction}{1}{%
  \pgfmathparse{%
    (and(   1,    #1<-1)*(1)            +%
    (and(#1>=-1,  #1< 1)*(-#1)   +%
    (and(#1>= 1,    1  )*(-1)%
    }%
}

\pgfmathdeclarefunction{SmoothFriction}{1}{%
  \pgfmathparse{%
    (and(   1,      1  )*(-tanh(#1) - 0.25 * #1 / ((0.25 * #1 * #1 + 0.75) * (0.25 * #1 * #1 + 0.75)))
    }%
}

\pgfmathdeclarefunction{TruePenetration}{1}{%
  \pgfmathparse{%
    (and(   1,    #1<0)*(100)            +%
    (and(#1>= 0,    1  )*(0)%
    }%
}

\pgfmathdeclarefunction{LinPenetration}{1}{%
  \pgfmathparse{%
    (and(     1,  #1< 0)*(-#1)   +%
    (and(#1>= 0,    1  )*(0)%
    }%
}

\pgfmathdeclarefunction{SmoothPenetration}{1}{%
  \pgfmathparse{%
    (and(     1,  #1< 0)*(#1*#1)   +%
    (and(#1>= 0,    1  )*(0)%
    }%
}

\begin{tikzpicture}
    \begin{axis}[
        name=plot-a,
        title=(a) Non-penetration,
        xmin = -2, xmax = 4,
        ymin = -0.3, ymax = 2,
        ylabel=Normal force $\lambda_n$,
        xlabel=Penetration distance $q_n$,
        xtick={0},
        ytick={0},
        legend pos  = north east,
        legend cell align={left},
        legend style={fill=white, fill opacity=0.6, draw opacity=1,text opacity=1, font=\small},
        grid=major,
        major grid style={line width=.1pt,draw=gray!20},
        width = 0.45\textwidth,
        height = 0.38\textwidth,       
    ]
    
        \addplot[domain=-2:4, blue, samples=500, ultra thick] {TruePenetration(x)};
        \addplot[domain=-2:4, red, samples=500, ultra thick] {LinPenetration(x)};
        \addplot[domain=-2:4, green, samples=500, ultra thick] {SmoothPenetration(x)};
        
        \legend{
            True,
            Linear~\eqref{eqn:lin-penetration},
            Smooth~\cite{heiden-2019-interactive-simulation}
        }
    \end{axis}
    \begin{axis}[
        name=plot-b,
        at={(plot-a.north east)},
        xshift=2.5cm,
        anchor=north west,
        title=(b) Friction,
        xmin = -5, xmax = 5,
        ylabel=Frictional force $\norm{\lambda_t}$,
        xlabel=Tangential velocity $\norm{\dot{q}_t}$,
        xtick={0},
        ytick={-1,0,1},
        yticklabels={$-\mu\lambda_n$,$0$,$\mu\lambda_n$},
        legend pos  = north east,
        legend cell align={left},
        legend style={fill=white, fill opacity=0.6, draw opacity=1,text opacity=1, font=\small},
        grid=major,
        major grid style={line width=.1pt,draw=gray!20},
        width = 0.45\textwidth,
        height = 0.38\textwidth,               
    ]
  
    \addplot[domain=-5:5, blue, samples=500, ultra thick] {TrueFriction(x)};
    \addplot[domain=-5:5, red, samples=500, ultra thick] {LinFriction(x)};
    \addplot[domain=-5:5, green, samples=500, ultra thick] {SmoothFriction(x)};

    \legend{
        True~\eqref{eqn:friction-dir},
        Linear~\eqref{eqn:lin-friction},
        Smooth~\cite{heiden-2021-neural-networks}
    }
  \end{axis}
\end{tikzpicture}
\caption{Comparison of different contact model implementations. The non-penetration requirements and Coulomb's friction law do not have well-defined gradients at $q_n=0$ and $\norm{\dot{q}_t}=0$, respectively. Compliant models relax these models by approximating the discontinuities, which we can consider as impulse function-like gradients, using functions with finite but sufficiently large gradients.}
\label{fig:contacts}
\end{figure*}
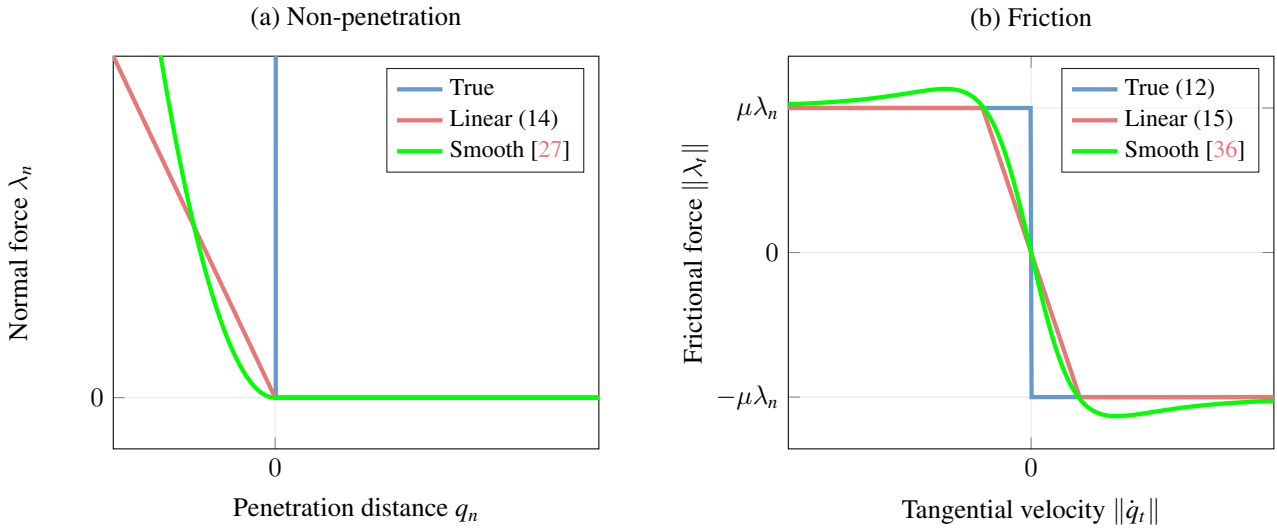


Complementarity-based contact models simulate ``hard'' contacts, \ie, there is strictly zero penetration between objects. Another way to model the contact physics is to approximate the contacts as soft contacts, \ie, some non-zero penetration between objects is allowed. Compliant models relax the step-like feature used for contact modeling (see \cref{fig:contacts}) functions by approximating them with finite but sufficiently large gradients. This allows gradients to be obtained directly from these functional approximations of non-penetration and friction forces. The tradeoff is that contact models are no longer strictly enforced, \eg, penetration is now modeled as a soft contact rather than a hard contact. The use of soft contacts can result in objects penetrating each other, however, this is normally aimed to be minimized.


The simplest approximation of non-penetration forces is to use a piece-wise linear function, \ie
\begin{equation}\label{eqn:lin-penetration}
    \lambda_n = k_n\max(-q_n, 0),
\end{equation}
where $k_n\in\mathbb{R}_+$ represents the gradient of the linearization, and $q_n\in\mathbb{R}$ is the penetration distance between two objects (negative means they are penetrating). This can be interpreted as modeling non-penetration as a spring-like force, or as using a ReLU activation function. This linearization is used by~\cite{geilinger-2020-add-contact,du-2021-diffpd-dynamics,DiffXPBD-stuyck-2023-dynamics}. A more complicated contact model for fabrics is used in~\cite{gong-2022-fine-fabrics}, which wraps a combination of bending and stretching forces inside a ReLU function. In~\cite{le-2023-differentiable-objects} where objects are modeled as density fields, the non-penetration force is proportionate to the volume of interpenetration between the objects. A potential issue with such linearizations is that there is zero gradient when contacts are inactive, making it difficult for gradient based optimizers to make new contacts~\cite{turpin-2022-grasp-hands}. In~\cite{turpin-2022-grasp-hands,turpin-2023-fast-simulation}, this is resolved by introducing a small gradient to the inactive portion of the compliant model, analogous to the leaky ReLU function. Apart from linear models of non-penetration forces, a quadratic approximation is used instead in~\cite{heiden-2019-interactive-simulation}, \ie, $\lambda_n=k_n\max(-q_n, 0)^2$. Some works use a more complicated spring-damper model for normal forces, where $\dot{q}_n$ is also considered in the model~\cite{heiden-2021-neural-networks,xu-2021-accelerated-simulation,jatavallabhula-2021-gradsim-control}.

For friction, Coulomb's law is commonly approximated as a piece-wise linear function, \ie
\begin{equation}\label{eqn:lin-friction}
    \lambda_t = - \frac{\dot{q}_t}{\norm{\dot{q}_t}_2} \min(\mu \lambda_n, k_t{\norm{\dot{q}_t}_2}),
\end{equation}
where $k_t\in\mathbb{R}_+$ determines the gradient of the linearization. This formulation is used by~\cite{geilinger-2020-add-contact,xu-2021-accelerated-simulation,turpin-2022-grasp-hands,du-2021-diffpd-dynamics,jatavallabhula-2021-gradsim-control}. More complicated models typically try to use a smoother approximation or model more complex dynamics not considered in Coulomb's friction law (\eg, modeling different static and dynamic friction coefficients)~\cite{heiden-2019-interactive-simulation,heiden-2021-neural-networks,gong-2022-fine-fabrics}.

\subsubsection{Position-based models}

Whereas complementarity-based and compliant models determine the effect of contacts on forces and velocities, another method is to directly constrain the positions of objects to avoid penetrations. 
A commonly-used framework to model this kind of contact model is XPBD (see Section~\ref{subsubsec:soft}) which is used by both Warp~\cite{macklin-2022-warp-graphics} and Brax~\cite{freeman-2021-brax-body}. 
    
Other position-based models first perform an update step without considering contact physics and then perform a correction step to satisfy contact-based constraints. \citet{turpin-2023-fast-simulation} implements position-based dynamics based on~\cite{deul-2016-position-dynamics}, where the correction step is performed using a Gauss-Seidel method. Coulomb's friction can also be accounted for using this approach, as discussed in~\cite{deul-2016-position-dynamics}. Alternatively, for cloth simulation, the correction step can be performed by using a quadratic program which minimizes the change in position of cloth particles while satisfying non-penetration constraints~\citep{qiao-2020-scalable-control,liang-2019-differentiable-problems}. Gradients are obtained from this optimization problem by using implicit differentiation on the KKT conditions. Both of these works use a QR decomposition trick to speed up the computation of gradients, taking advantage of the fact that the number of non-penetration constraints is typically much smaller than the number of simulation variables.

\subsubsection{MLS-MPM}
For the MLS-MPM dynamics model (see Section~\ref{subsubsec:soft}), which incorporates both rigid-body and soft-body objects, there are two types of contacts that are considered. First, self-collisions of soft-body objects are handled ``automatically'' by the MLS-MPM model, which accounts for momentum transfer between neighboring soft-body particles and grids. Second, collisions between soft-bodies and rigid-body obstacles are treated as boundary conditions, which are similar to the complementarity conditions given by~\cref{eqn:normal,eqn:friction-cone}, and enforced by simply projecting the velocities into these sets after they have been updated. Gradients for these collision models are derived analytically in~\cite{hu-2019-chainqueen-robotics,spielberg-2023-advanced-chainqueen} to make the differentiable simulator as high-performance as possible.

\subsection{Integrator}
\label{subsec:integrator}





An integration scheme is required to use the computed forces to propagate a given dynamic model forwards in time. Throughout this section, the notation $q_k,\dot{q}_k,\ddot{q}_k\in\mathbb{R}^n$ represents the position, velocity, and acceleration state vectors at the $k$-th time step, respectively, and $\Delta t\in\mathbb{R}_+$ is the time step between each iteration.

Two main categories of integration schemes are explicit (Section~\ref{subsubsec:explicit}) and implicit (Section~\ref{subsubsec:implicit}) integration methods. Although explicit integration is easier to implement in practice, it is well known that it is less numerically stable compared to implicit integration schemes. For example, experiments in~\cite{du-2021-diffpd-dynamics} show that implicit integration is stable at time steps of up to $10\textnormal{ms}$, whereas explicit integration is only stable at time steps of up to $0.5\textnormal{ms}$. A consequence of this is that explicit schemes have a larger memory overhead due to needing to simulate more time steps, and therefore requires tricks such as ``checkpointing'' to reduce this memory overhead~\cite{qiao-2021-efficient-bodies,spielberg-2023-advanced-chainqueen,hu-2020-difftaichi-simulation}. Although explicit integration schemes are relatively straightforward to differentiate through, implicit integration schemes, which involve solving a nonlinear system of equations, require more complex techniques to extract gradients from. For both techniques, a standard way to improve the stability and accuracy is to use higher order integration methods, such as the Runge–Kutta family of integration schemes. We refer to, \eg,~\cite{volino2001comparing} for a more in depth comparison of these integration methods.

\subsubsection{Explicit integration}\label{subsubsec:explicit}

The simplest integrator, explicit (forwards) Euler integration updates an object's position and velocity by taking a small time step and applying the current velocity to the position and the current acceleration to the velocity, \ie
\begin{subequations}\label{eqn:explicit-int}
    \begin{align}
        \dot{q}_{k+1} &= \dot{q}_k + \Delta t \cdot \ddot{q}_k \\
        q_{k+1} &= q_k + \Delta t \cdot \dot{q}_k.
    \end{align}
\end{subequations}
This method is used by~\cite{de-2018-end-control,hu-2019-chainqueen-robotics,werling-2021-fast-contact,qiao-2021-efficient-bodies, chen-2023-daxbench-physics, granados-2022-model-physics, spielberg-2023-advanced-chainqueen, huang-2021-plasticinelab-physics, song-2020-learning-simulations, wang-2022-differentiable-musculotendons, wiedemann-2023-training-gradient,le-2021-differentiable-identification, liu-2023-robotic-dynamics, holl-2020-phiflow-simulations}. Explicit integration schemes are desirable as it is straightforward to propagate gradients through time just by using the chain rule. However, it is well-known that explicit integration is conditionally stable~\cite{baraff-1998-large-simulation,fang-2018-a-stepping,macklin-2019-small-simulation}, \ie, there exists a critical time step $t_\mathrm{crit} > 0$ such that the simulation can be unstable if $\Delta t \geq t_\mathrm{crit}$. This can lead to stability and accuracy issues, particularly with more complex simulations.

To obtain better stability properties, higher-order explicit schemes are used in~\cite{bezgin-2023-JAX-flows,petrik-2022-learning-physics,zhong-2021-extending-models, zhao-2022-automatic-grammar}, \eg the explicit fourth-order Runge-Kutta method (RK4), which offers improved accuracy and stability compared to the basic Euler method by using multiple stages to update positions and velocities~\citep{press-1992-numerical-computing}. Other works~\cite{degrave-2019-a-robotics,heiden-2021-neural-networks,freeman-2021-brax-body,gartner-2022-differentiable-reconstruction,jatavallabhula-2021-gradsim-control,heiden-2021-disect-cutting,heiden-2019-interactive-simulation,xu-2021-accelerated-simulation,turpin-2022-grasp-hands,turpin-2023-fast-simulation, wang-2021-sim2sim-robots, heiden-2019-real2sim-physics, wang-2023-softzoo-environments} instead use semi-implicit Euler integration (symplectic integration), with the following steps
\begin{subequations}\label{eqn:semi-implicit-int}
    \begin{align}
        \dot{q}_{k+1} &= \dot{q}_k + \Delta t \cdot \ddot{q}_k \\
        q_{k+1} &= q_k + \Delta t \cdot \dot{q}_{k+1} ,
    \end{align}
\end{subequations}
\ie the velocity is first updated explicitly, then position is updated using this new velocity $\dot{q}_{k+1}$, rather than the current velocity $\dot{q}_k$, as in explicit integration. These semi-implicit steps are similarly easy to implement and backpropagate gradients through as explicit steps. 

An issue shared by these methods is that, as discussed in~\citep{qiao-2021-efficient-bodies}, to backpropagate gradients through time, information at every time step needs to be stored in memory. This can be prohibitively expensive when the total simulation horizon is large, particularly as explicit integration schemes require small time steps to be stable. In~\cite{qiao-2021-efficient-bodies,spielberg-2023-advanced-chainqueen,hu-2020-difftaichi-simulation}, checkpointing methods are proposed to reduce memory requirements of the simulator in exchange for slower runtime. During the forwards pass, information is ``checkpointed'' rather than storing the entire computational graph (\eg, only store the simulation state for every $n$ time steps). In the backwards pass, information required to compute the gradients that was lost can be recovered by re-running the simulation from the closest checkpoint.

\subsubsection{Implicit integration}\label{subsubsec:implicit}

Implicit integration is designed to handle stiff differential equations and constraints more effectively than explicit methods, and therefore allows for much larger time steps to be used~\cite{baraff-1998-large-simulation,fang-2018-a-stepping,macklin-2019-small-simulation}. Implicit (backwards) Euler integration computes the state at the next time step by using the gradient at the next time step, \ie
\begin{subequations}\label{eqn:implicit-int}
    \begin{align}
        \dot{q}_{k+1} &= \dot{q}_k +  \Delta t \cdot \ddot{q}_{k+1}   \\
        q_{k+1} &= q_k +  \Delta t \cdot \dot{q}_{k+1}.
    \end{align}
\end{subequations}
The implicit Euler method is used by~\cite{geilinger-2020-add-contact,liang-2019-differentiable-problems,qiao-2020-scalable-control,gong-2022-fine-fabrics,du-2021-diffpd-dynamics,li-2022-diffcloth-contact,du-2021-underwater-simulation,DiffXPBD-stuyck-2023-dynamics, holl-2020-phiflow-simulations}. Other first order implicit methods have been used, such as \cite{wang-2022-differentiable-musculotendons} which uses a first-order backwards differentiation formula (BDF1). Like explicit methods, higher-order implicit methods can be used to improve stability and accuracy. For example, the second-order backwards differentiation formula (BDF2) is used in~\cite{mora-2021-pods-simulation,Xu-2021-An-Design}, and the implicit second-order Runge-Kutta (RK2) is used in~\cite{ding-2021-dynamic-language,wang-2022-differentiable-musculotendons}. In~\cite{wang-2022-recurrent-data}, only some of the state variables are updated using implicit integration, while the rest are updated using explicit integration, to reduce the computational cost of solving a large system of equations. In~\cite{le-2023-differentiable-objects,howelll-2022-dojo-robotics}, a variational integration scheme is used, where instead of discretizing the system dynamics as in~\cref{eqn:implicit-int}, the integration scheme is derived by discretizing Hamilton's Principle. This results in an integration scheme which conserves momentum and energy, and is therefore stable even for relatively large simulation time steps. 

Unlike explicit Euler, implicit Euler steps represent a set of (possibly) non-linear equations which must be solved to find the state at the next time step. A standard way to solve these equations is to use an iterative method, such as Newton's root-finding method~\cite{howelll-2022-dojo-robotics,Xu-2021-An-Design,geilinger-2020-add-contact}. Alternatively, some works take a first-order Taylor-series expansion of the equations~\cite{liang-2019-differentiable-problems,qiao-2020-scalable-control,gong-2022-fine-fabrics}. In this case, solving for the state at the next time step involves a simple factor-solve of the resulting linear system of equations. For both of these methods, the gradient can be computed by implicitly differentiating through the equations (see Section~\ref{subsubsec:analytical}). The adjoint method is commonly used here to make the gradient computation more efficient (see Section~\ref{subsubsec:analytical}). 

Alternatively, in~\cite{du-2021-diffpd-dynamics,li-2022-diffcloth-contact}, Projective Dynamics (see~\cref{subsubsec:soft}) is used to express the nonlinear set of equations~\cref{eqn:implicit-int} as an energy minimization problem. By making certain assumptions on the system dynamics, this optimization problem can be solved by using an alternating optimization algorithm. These steps can be cheaper than a na\"ive Newton's method by taking advantage of the fact that the Cholesky factorization of one of the matrices arising from the alternating optimization scheme can be precomputed. The gradient backpropagation step, which is still derived by implicitly differentiating the optimality conditions, can also take advantage of the same precomputed Cholesky factorization. Numerical experiments in~\cite{du-2021-diffpd-dynamics} demonstrate the computational superiority of the Projective Dynamics approach compared to na\"ive implicit differentiation implementations. 
\section{Differentiable Simulators}
\label{sec:differentiable_simulators}

\begin{table*}[]
\rowcolors{2}{gray!30}{white}
\centering
\caption{A table featuring eleven open-source differentiable simulators. Eight of these simulators have been utilized in existing literature reviewed herein, while the remaining three (below the bolded line) have not yet been adopted due to their recent introduction. This is a subset of the engines found in \cref{tab:background}, with a focus on user features researches may consider when choosing a differentiable simulator. All engines have Python bindings and can be run on GPU resources. There are three types of objects supported: Rigid, Deformable (Def) which includes cloth and Fluids.}\label{tab:engines}
\resizebox{0.99\textwidth}{!}{%
\begin{tabular}{|l|ccc|c|cc|ccc|c|cccc|c|}
\hline
                              & \multicolumn{3}{c|}{Object Types} &          & \multicolumn{2}{c|}{Integrations} & \multicolumn{3}{l|}{File Types Supported}    &                    & \multicolumn{4}{c|}{Collision Type}       &                \\ \hline
Engine                       & Rigid                & Def  & Fluids         & Language & Jax             & PyTorch         & URDF       & MJCF       & USD        & \makecell[c]{Parallel\\Sim}  & Primitive  & Mesh       & Particle   & SDF        & \makecell[c]{Diff\\Render} \\ \hline
TDS~\cite{heiden-2019-interactive-simulation} & \ForestGreencheck           &          &            & C++      &                 &                 & \ForestGreencheck &            &            & \ForestGreencheck         & \ForestGreencheck &            &            &            &                \\
Warp~\cite{macklin-2022-warp-graphics}                          & \ForestGreencheck           & \ForestGreencheck  &        & Python   & \ForestGreencheck      & \ForestGreencheck      & \ForestGreencheck & \ForestGreencheck           & \ForestGreencheck & \ForestGreencheck         & \ForestGreencheck & \ForestGreencheck & \ForestGreencheck & \ForestGreencheck & \ForestGreencheck               \\
DiffTaichi~\cite{hu-2020-difftaichi-simulation}                    & \ForestGreencheck           & \ForestGreencheck       &   & C++      & \ForestGreencheck      & \ForestGreencheck      &            &            &            &                    &            &            & \ForestGreencheck &            &                \\
Brax~\cite{freeman-2021-brax-body}                          & \ForestGreencheck           &                    &  & Python   & \ForestGreencheck      &      & \ForestGreencheck & \ForestGreencheck &            & \ForestGreencheck         & \ForestGreencheck & \ForestGreencheck &            &            &                \\
Nimble~\cite{werling-2021-fast-contact}                        & \ForestGreencheck           &                  &    & C++      &                 & \ForestGreencheck      & \ForestGreencheck &            &            &                    & \ForestGreencheck & \ForestGreencheck &            &            &                \\
Dojo~\cite{howelll-2022-dojo-robotics}                          & \ForestGreencheck           &                     & & Julia    & \ForestGreencheck      & \ForestGreencheck      & \ForestGreencheck &            &            &                    & \ForestGreencheck &            &            &            &                \\
GradSim~\cite{jatavallabhula-2021-gradsim-control}                      & \ForestGreencheck           & \ForestGreencheck         &  & C++      &                 & \ForestGreencheck      & \ForestGreencheck &            & \ForestGreencheck &                    & \ForestGreencheck & \ForestGreencheck &            &            & \ForestGreencheck     \\
PhiFlow~\cite{holl-2020-phiflow-simulations} & & & \ForestGreencheck & Python & \ForestGreencheck & \ForestGreencheck & & & & & & & & & \\ \Xhline{4\arrayrulewidth}
daX~\cite{chen-2023-daxbench-physics}                           & \ForestGreencheck           & \ForestGreencheck        &   & Python   & \ForestGreencheck      & \ForestGreencheck      & \ForestGreencheck & \ForestGreencheck & \ForestGreencheck & \ForestGreencheck         & \ForestGreencheck & \ForestGreencheck &            &            & \\ 
FluidLab~\cite{xian-2023-fluidlab-manipulation} & \ForestGreencheck & \ForestGreencheck & \ForestGreencheck & Python & & \ForestGreencheck & & & & & \ForestGreencheck & \ForestGreencheck & \ForestGreencheck & \ForestGreencheck & \\
JAX-Fluids~\cite{bezgin-2023-JAX-flows} & & & \ForestGreencheck & Python & \ForestGreencheck & & & & & & & & & &\\ \hline
\end{tabular}}
\end{table*}
This section aims to be a more practical guide to differentiable simulation engines. Throughout the literature on differentiable simulators, eight differentiable simulators were identified~\cite{macklin-2022-warp-graphics,heiden-2021-neural-networks,werling-2021-fast-contact,hu-2020-difftaichi-simulation,freeman-2021-brax-body,chen-2023-daxbench-physics,holl-2020-phiflow-simulations,howelll-2022-dojo-robotics} that were adopted by other works included within the scope of this review. \cref{tab:engines} summarizes the key features of these eight engines and three additional differentiable physics engines. While the latter three have been recently proposed and not yet widely adopted, they show promise for practical applications. Most of the simulator codebases focus on rigid body simulation~\cite{macklin-2022-warp-graphics, heiden-2021-neural-networks, werling-2021-fast-contact, freeman-2021-brax-body, howelll-2022-dojo-robotics}, however, some do support soft body simulation~\cite{hu-2020-difftaichi-simulation, macklin-2022-warp-graphics} and fluid simulation~\cite{holl-2020-phiflow-simulations,bezgin-2023-JAX-flows, xian-2023-fluidlab-manipulation}. 


\subsection{Warp}

NVIDIA Warp~\cite{macklin-2022-warp-graphics} is a Python framework designed for high-performance simulation and graphics code, featuring just-in-time (JIT) compilation of Python functions for efficient CPU and GPU execution. 


Warp allows for both soft and rigid body simulation, as well as supporting two types of integrators, semi-implicit and XPBD~\cite{macklin-2016-xpbd-dynamics}. Importing can be done by adding joints manually or through importing either a URDF, MJCF or USD. It supports non-differentiable rendering and basic differentiable raycasting to visualize the model and is based on a maximal coordinate system. Warp supports triangular-based meshes which can be created through the API by providing points, indices and velocities. Most of the underlying functionality in Warp has featured in other Nvidia differentiable simulation engines, such as gradSim~\cite{jatavallabhula-2021-gradsim-control}, before being distilled as Warp.



\subsection{Tiny Differentiable Simulator}

Tiny Differentiable Simulator (TDS)~\cite{heiden-2021-neural-networks} is a C++ and CUDA physics library, known for its simplicity and independence from external dependencies. TDS focuses on rigid-body dynamics and contacts, therefore, TDS does not support deformable objects. A limitation of the simulator is its support for only primitive collision shapes, like spheres, planes and capsules. One key advantage of TDS is its ability to run thousands of parallel simulations on a single GPU. TDS does not have detailed documentation describing the API, however, it does have a number of comprehensive examples.

\subsection{Nimble}
Nimble~\cite{werling-2021-fast-contact}, stemming from a fork of the popular DART physics engine~\citep{lee-2018-dart-toolkit}, has evolved into a fast and fully differentiable physics engine with analytical gradients and PyTorch bindings. Nimble supports non-differentiable browser-based rendering of the simulation scene. However, Nimble was designed for single-threaded execution and does not support parallel environments.

\subsection{DiffTaichi}

DiffTaichi~\cite{hu-2020-difftaichi-simulation}, implemented using the Taichi~\cite{hu-2019-taichi-structures} differentiable programming framework, supports both rigid-body and soft-body simulations. Several standalone environments exist in DiffTaichi operating as examples of capability, however, unlike other simulators within this section DiffTaichi is not intended a general purpose physics engine, rather aiming to provide a method to calculate gradients of dynamical systems effectively and efficiently through source code transformations~\cite{griewank-2008-evaluating-algorithmic}. As such, DiffTaichi does not support common file types, such as URDF, MJCF or USD, and kinematic chains needs to be entered pragmatically via the API. 



\subsection{Brax}

Brax~\cite{freeman-2021-brax-body}, which is based on JAX~\cite{jax2018github}, allows for scalability across parallel simulations. Brax supports several backend physics pipelines and supports both MJCF and URDF file loading as well as programmatically creating object models. Brax supports a non-differentiable browser rendering system. 

\subsection{GradSim}

GradSim~\cite{jatavallabhula-2021-gradsim-control} ($\nabla$Sim) aims to unify the growing popularity of differentiable rendering~\cite{kato-2020-differentiable-survey} with differential physics. GradSim supports simulation of both rigid and deformable objects. However, GradSim does not support parallel execution or include documentation about the API, this could increase the learning curve for new users.



\subsection{Dojo}

Dojo~\cite{howelll-2022-dojo-robotics} is a differentiable physics engine tailored for robotics. Dojo uses a variational integrator for stability and to help with handling large timesteps. Dojo does not support mesh-based collisions, only allowing collisions between primitive shapes. Dojo is no longer actively maintained, however, there is evidence of ongoing adoption of this engine throughout the literature. 



\subsection{PhiFlow}

PhiFlow ($\phi$Flow)~\cite{holl-2020-phiflow-simulations} aims to create a differenitable simulation framework for solving partial differential equations (PDE), mostly focusing on fluid simulation applications. It supports both Pytorch~\cite{paszke-2017-automatic-pytorch} and JAX~\cite{jax2018github}, as well as a non-differentiable HTML-based rendering.  Objects can be loaded from \textit{su2}~\cite{economon-2016-su2-design} CFD files, which is an open source CFD simulation software.

\subsection{Additional Differentiable Simulators}

Some differentiable simulators have recently been proposed and although they have not been adopted by other research, they are worth noting. DaXBench~\cite{chen-2023-daxbench-physics} is one such simulator. DaXBench supports various deformable objects like fluids, ropes, and cloth. daX uses JAX~\cite{jax2018github} and integrates seamlessly with OpenAI Gym~\cite{brockman-2016-openai-gym}. DaXBench aims to serve as a tool for standardized testing and development of new approaches in deformable object manipulation using differentiable simulation. 

FluidLab~\cite{xian-2023-fluidlab-manipulation} is another such differentiable simulator and is tailored towards fluid manipulation task, but also supports rigid and deformable object manipulation. The differentiable simulator is called FluidEngine and is built using the Taichi~\cite{hu-2020-difftaichi-simulation} differentiable programming paradigm. The focus of FluidLab is $10$ distinct manipulation tasks, where the agent has to interact with certain types of fluids to achieve a goal. 

JAX-Fluids~\cite{bezgin-2023-JAX-flows} is the final noteworthy inclusion. JAX-Fluids is written in JAX~\cite{jax2018github} and designed for CFD. It incorporates state-of-the-art numerical methods and is intended to allow integration of CFD within machine learning workflows. JAX-Fluids aims to handle complex fluid dynamics scenarios, including three-dimensional turbulence, compressibility effects, and two-phase flows. The configurations for the simulator rely on the JSON format but loading objects within the simulator is currently not supported. The simulator does not currently support parallel simulations, however, the authors have stated their intention to support this capability in the near future. 



\section{Applications}

In the literature, differentiable simulators have been applied as a solution for many problems. These applications of differentiable simulators all fundamentally involve solving an optimization problem, which are typically solved using gradient-based optimization algorithms. For an overview of these optimization algorithms, we refer the reader to \cite{ruder2016overview}.

\begin{figure*}[h!]
    \includegraphics[width=1\linewidth]{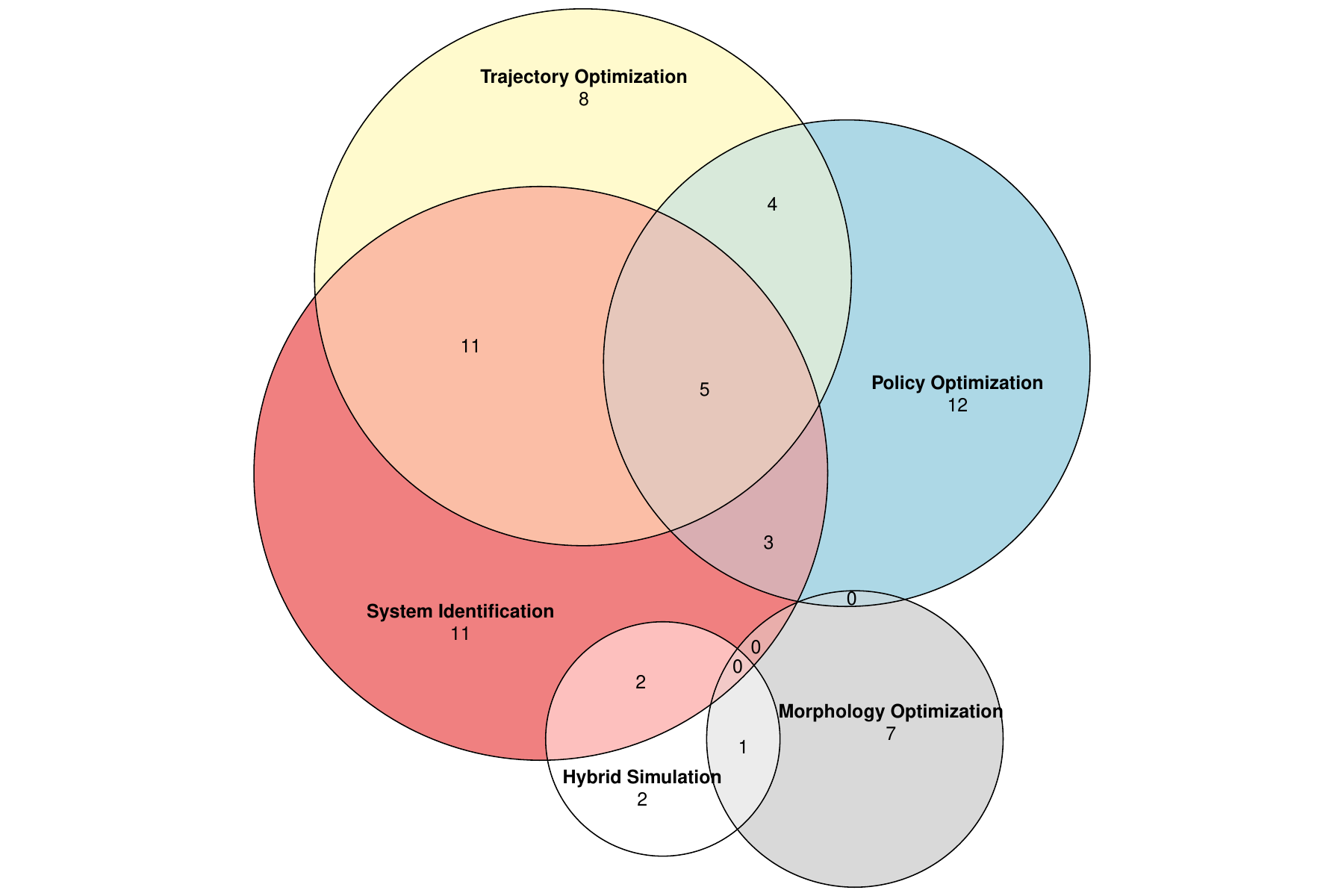}
    \caption{An Area-Proportional Venn diagram~\cite{larsson-2018-case-eulerr} visualizing the different application areas of differentiable simulators explored in this review with the numbers indicating the number of referenced works. The number of cited works for a given application area in the Venn diagram is cumulative.}
    \label{fig:app-venn-diagram}
\end{figure*}

We review five primary types of application domains:
\begin{enumerate}
    \item System Identification -- leverages differentiable simulators to model complex systems, allowing for efficient parameter tuning and system characterization.
    \item Trajectory Optimization -- employs differentiability to refine and optimize the path of objects or entities within a simulated environment, enhancing control capabilities.
    \item Morphological Optimization -- utilizes differentiable simulators to evolve and optimize the physical structures or morphologies of entities. Typically, morphology optimization is undertaken as co-optimization of both morphology and control.
    \item Policy Optimization -- uses differentiable simulators to train and optimize policies, particularly in a reinforcement learning framework, facilitating the development of robust and adaptive decision-making strategies.
    \item Neural Network Augmented Simulation - integrates neural networks within the simulator to closer align simulations to the real-world.
\end{enumerate}
The application areas of differentiable simulators listed above are categorized based on the typical distinctions made in literature. All papers that investigate use cases of differentiable simulators can be categorized into one of these five application areas. The number and overlap of works in each application domain are summarized in \cref{fig:app-venn-diagram}. 




\label{sec:applications}


\subsection{System Identification}
\label{sec:sysID}
System identification aims to construct an accurate mathematical model of a physical system by estimating unknown parameters based on observations of the system. To estimate the model parameters $\theta\in\mathbb{R}^p$, most works minimize the mean squared error between the observed state trajectory $q_1,\ldots,q_N \in \mathbb{R}^n$, and the state trajectory estimated by the physics simulator $\hat{q}_1(\theta),\ldots,\hat{q}_N(\theta) \in \mathbb{R}^n$ given the model parameters $\theta$, \ie,
\begin{subequations} \label{eqn:sysid}
    \begin{align}
        \minimize_{\theta} \quad & \sum_{i=1}^N~\norm{\hat{q}_i(\theta) - q_i}_2^2  \label{eqn:sysid-obj}\\
        \subjto \quad & \theta_{min} \leq \theta \leq \theta_{max},
    \end{align}
\end{subequations}
where $\theta_{min},\theta_{max}\in\mathbb{R}^p$ are the minimum and maximum allowable values of the parameters. 

\begin{figure}
    \centering
    \begin{subfigure}{\linewidth}
		\includegraphics[width=\linewidth]{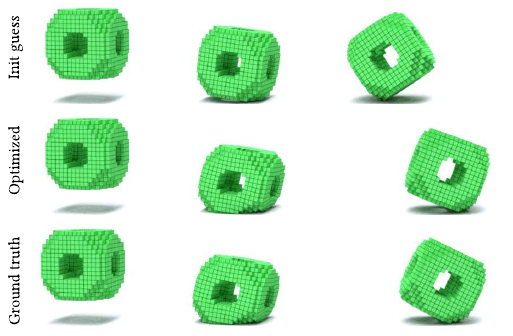}
        \caption{System identification of a simulated bouncing ball~\citep{du-2021-diffpd-dynamics}. Shown are the initial frames (left), the moment the ball collides with the ground (middle), and final frames (right) of trajectories arising from randomized (top), optimized (middle), and ground truth (bottom) material parameters.}
        \label{fig:sys-id-diffpd}
    \end{subfigure}\vspace{0.5cm}\\
    \begin{subfigure}{\linewidth}
		\includegraphics[width=\linewidth]{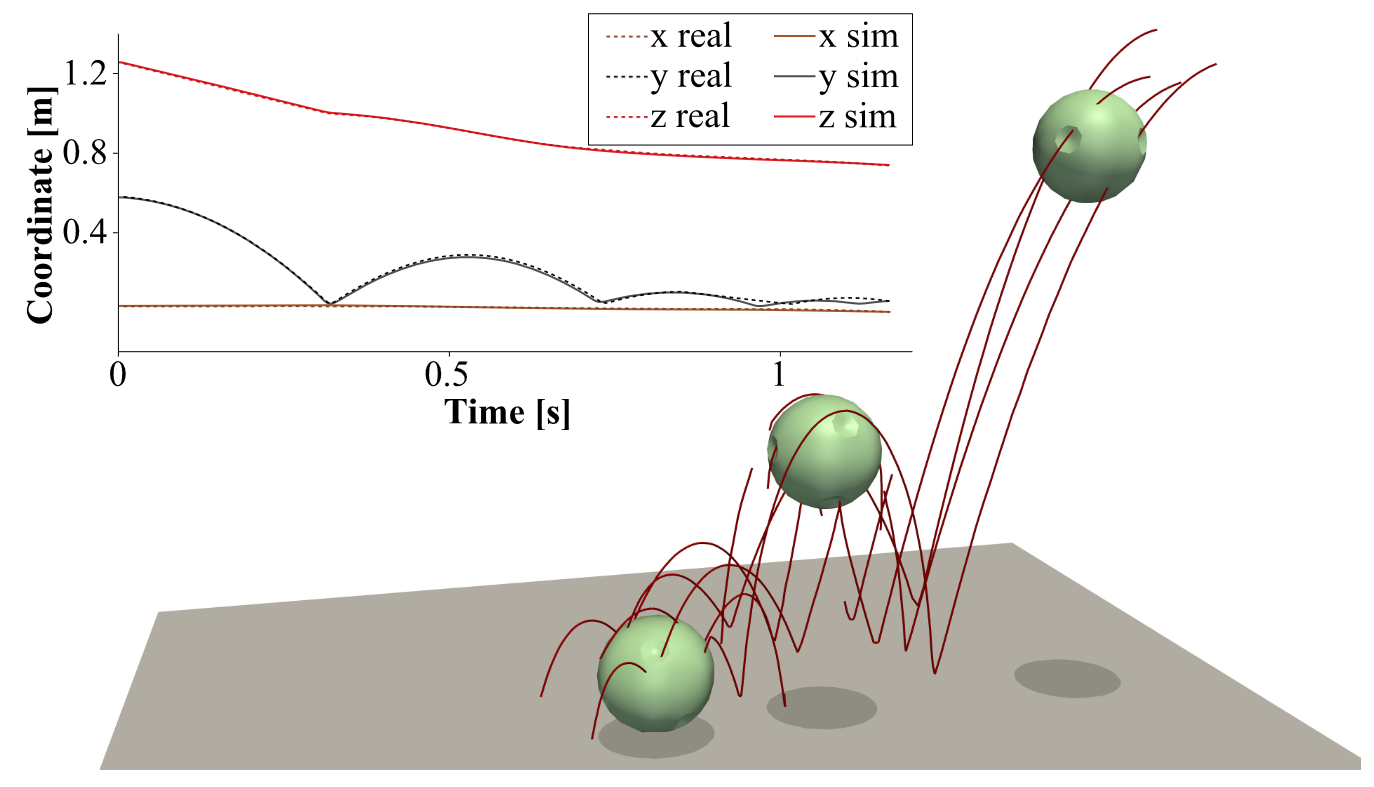}
        \caption{System identification of a real-world bouncing ball~\citep{geilinger-2020-add-contact}. The red lines show the motion-captured trajectories of the ball, and the green ball shows the estimated trajectory of the ball with optimized material parameters. The graph compares the real-world and estimated trajectories of the ball.}
        \label{fig:sys-id-add}
    \end{subfigure}
    \caption{Examples of applications of differentiable simulation to perform system identification. Both examples estimate the material parameters (\eg, stiffness and dampening ratio) of an elastic bouncing ball so that the trajectory of the ball matches the ground truth as closely as possible.}
\end{figure}

System identification has been used to estimate the inertial properties (\eg, mass and moment of inertia) of colliding objects or objects given an initial known impulse~\citep{de-2018-end-control,hu-2019-chainqueen-robotics,qiao-2020-scalable-control,jatavallabhula-2021-gradsim-control,ding-2021-dynamic-language}. Similarly, inertial and kinematic properties (\eg, length) of a pendulum-like system have been estimated in~\cite{heiden-2019-real2sim-physics,lutter-2021-differentiable-learning,Heiden2022InferringVideo}. Other works estimate the stiffness and elasticity properties (\eg, Young's modulus, Poisson's ratio) of cloth~\citep{liang-2019-differentiable-problems,li-2022-diffcloth-contact,gong-2022-fine-fabrics,sundaresan-2022-diffcloud-objects}, bouncing objects~\citep{du-2021-diffpd-dynamics,zhong-2021-extending-models} (see \cref{fig:sys-id-diffpd}), rope~\citep{liu-2023-robotic-dynamics}, or actuated soft robots~\citep{spielberg-2023-advanced-chainqueen,dubied-2022-simtoreal-actuation,arnavaz-2023-differentiable-simulations}. Other common applications include estimating the coefficients of friction of robots~\citep{granados-2022-model-physics,wang-2022-real2sim2real-engine} or other objects sliding on a surface~\citep{le-2021-differentiable-identification,le-2023-differentiable-objects,song-2020-learning-simulations}, and to identify a recorded trajectory of a robot~\cite{petrik-2022-learning-physics, li-2022-d&d-camera} or human~\cite{gartner-2022-differentiable-reconstruction}. In~\cite{heiden-2021-disect-cutting}, it was shown how differentiable simulators could be used to estimate properties of a knife and a soft object that it was cutting, as well as the cutting force being applied. In~\cite{lv-2022-samrl-rendering,lv-2022-sagci-learning}, URDF models of household objects which are initially obtained from a scanned point cloud are iteratively improved using differentiable simulation, by using trajectories observed from a robot agent interacting with the object. Some works perform system identification using differentiable rendering techniques~\citep{jatavallabhula-2021-gradsim-control,arnavaz-2023-differentiable-simulations,lv-2022-samrl-rendering,petrik-2022-learning-physics,sundaresan-2022-diffcloud-objects,Heiden2022InferringVideo}, where the loss function \cref{eqn:sysid-obj} is instead defined as the pixelwise mean square error between a recorded video of the true system, and a rendered video from the differentiable simulator. Many works apply system identification to estimate parameters of real-world systems~\citep{du-2021-underwater-simulation,le-2021-differentiable-identification,lutter-2021-differentiable-learning,li-2022-diffcloth-contact,sundaresan-2022-diffcloud-objects,le-2023-differentiable-objects,liu-2023-robotic-dynamics,granados-2022-model-physics,wang-2022-real2sim2real-engine,chen-2022-real-simulation,arnavaz-2023-differentiable-simulations,heiden-2019-real2sim-physics,heiden-2021-disect-cutting,ding-2021-dynamic-language,song-2020-learning-simulations,lv-2022-samrl-rendering,collins-2021-follow-realitygrad,lv-2022-sagci-learning,petrik-2022-learning-physics,Heiden2022InferringVideo}.

\textbf{Advantages and Challenges: } 
Compared to black-box or gradient-free methods for system identification, the use of differentiable simulations has been shown to converge to lower cost solutions in less computational time~\citep{liang-2019-differentiable-problems,jatavallabhula-2021-gradsim-control,li-2022-diffcloth-contact,zhong-2021-extending-models,sundaresan-2022-diffcloud-objects,granados-2022-model-physics}. \citet{gong-2022-fine-fabrics} showed that differentiable simulators enabled accurate system identification of cloth parameters by only using a small amount of recorded time steps, which only contains subtle movements of the cloth. This is as opposed to other approaches which were unable to accurately estimate the cloth parameters using the same limited set of data. In~\citep{du-2021-diffpd-dynamics,le-2021-differentiable-identification}, it was noted that by nature of the formulation of the optimization problem \cref{eqn:sysid}, it is possible to obtain parameters which are quite different from the ground truth values, while still yielding similar predicted trajectories. This problem is called ``parameter observability'' in~\citep{le-2021-differentiable-identification}. The authors show how this is a larger issue when there are more parameters that are estimated simultaneously, and how data of more complex trajectories (\eg, collisions with objects with known parameters) can help to circumvent this issue by exposing certain desired parameters of the system.

\textbf{System Identification and Control: } 
One of the main uses of system identification is for the estimated parameters to be used for control tasks~\citep{du-2021-underwater-simulation,lutter-2021-differentiable-learning,granados-2022-model-physics,chen-2022-real-simulation,wang-2022-real2sim2real-engine,wang-2021-sim2sim-robots,song-2020-learning-simulations,lv-2022-samrl-rendering,collins-2021-follow-realitygrad,lv-2022-sagci-learning,Heiden2022InferringVideo}, \ie, for use with trajectory optimization (see \cref{Trajectory Optimization}), policy optimization (see \cref{subsec:policy-opt}), or with more classical control methods. For example, in~\citep{lutter-2021-differentiable-learning} it was shown how system identification allowed a robot to learn how to perform the complex ``ball-in-cup'' task, whereas black-box methods failed to perform the same task. In this control context, system identification has also been performed in an online fashion, where system identification is performed on continuously measured trajectory data while an agent is simultaneously being controlled to perform a task. This has been shown to be advantageous for dynamically changing environments~\citep{granados-2022-model-physics,chen-2022-real-simulation}, to continuously improve the accuracy of the estimated parameters~\citep{wang-2022-real2sim2real-engine}, or to avoid the need for collecting offline data beforehand~\citep{du-2021-underwater-simulation}. In~\citep{chen-2022-real-simulation}, a trajectory buffer system was used to account for noisy and outdated observations, which would otherwise lead to inaccurate or unstable parameters estimations.


\subsection{Trajectory Optimization} \label{Trajectory Optimization}
\begin{figure}[tbp]
    \centering
    \includegraphics[width=0.4\linewidth, trim=20cm 0 25cm 4cm, clip]{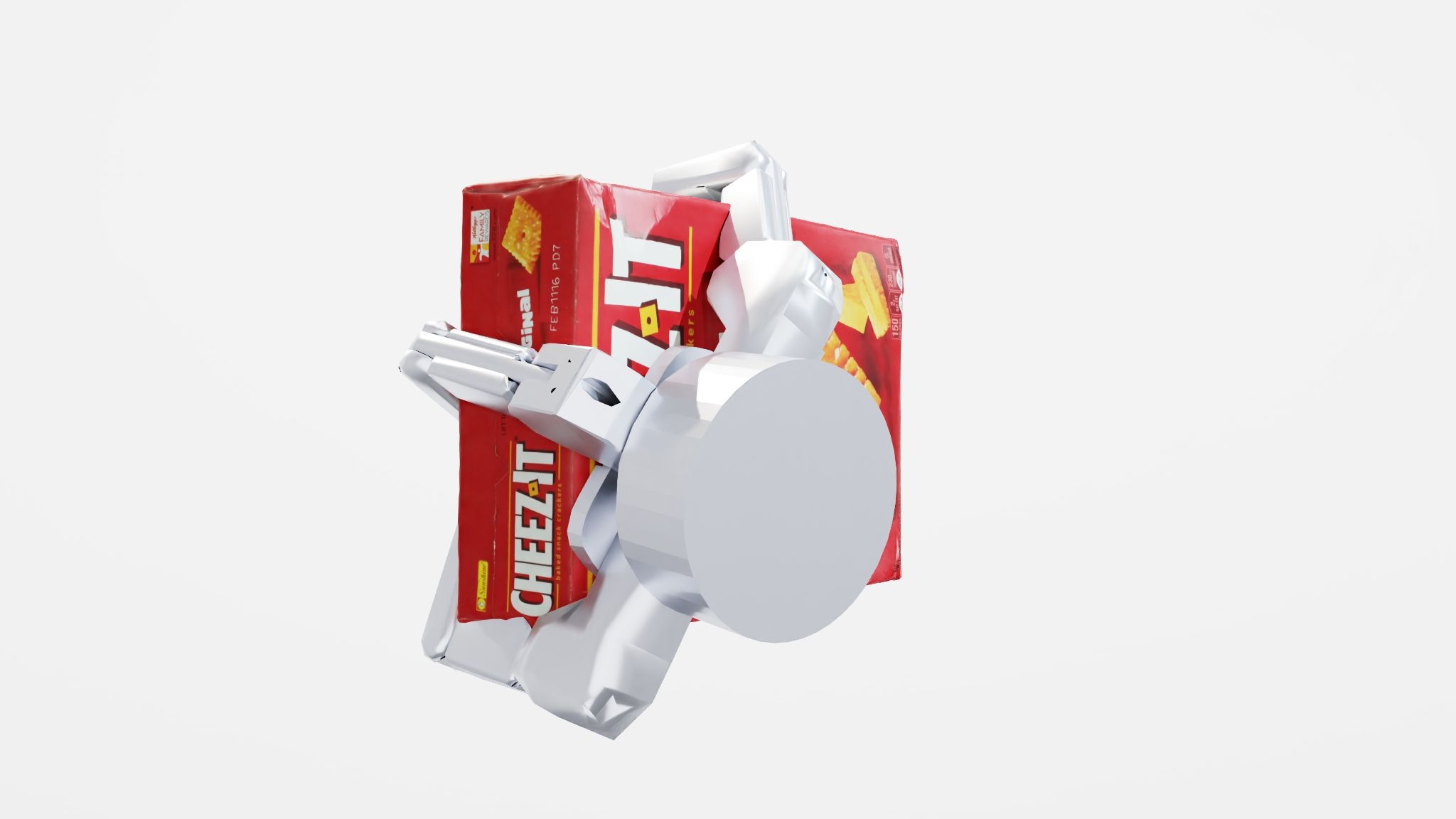}%
    \includegraphics[width=0.5\linewidth, trim=0cm 0cm 0cm 35cm, clip]{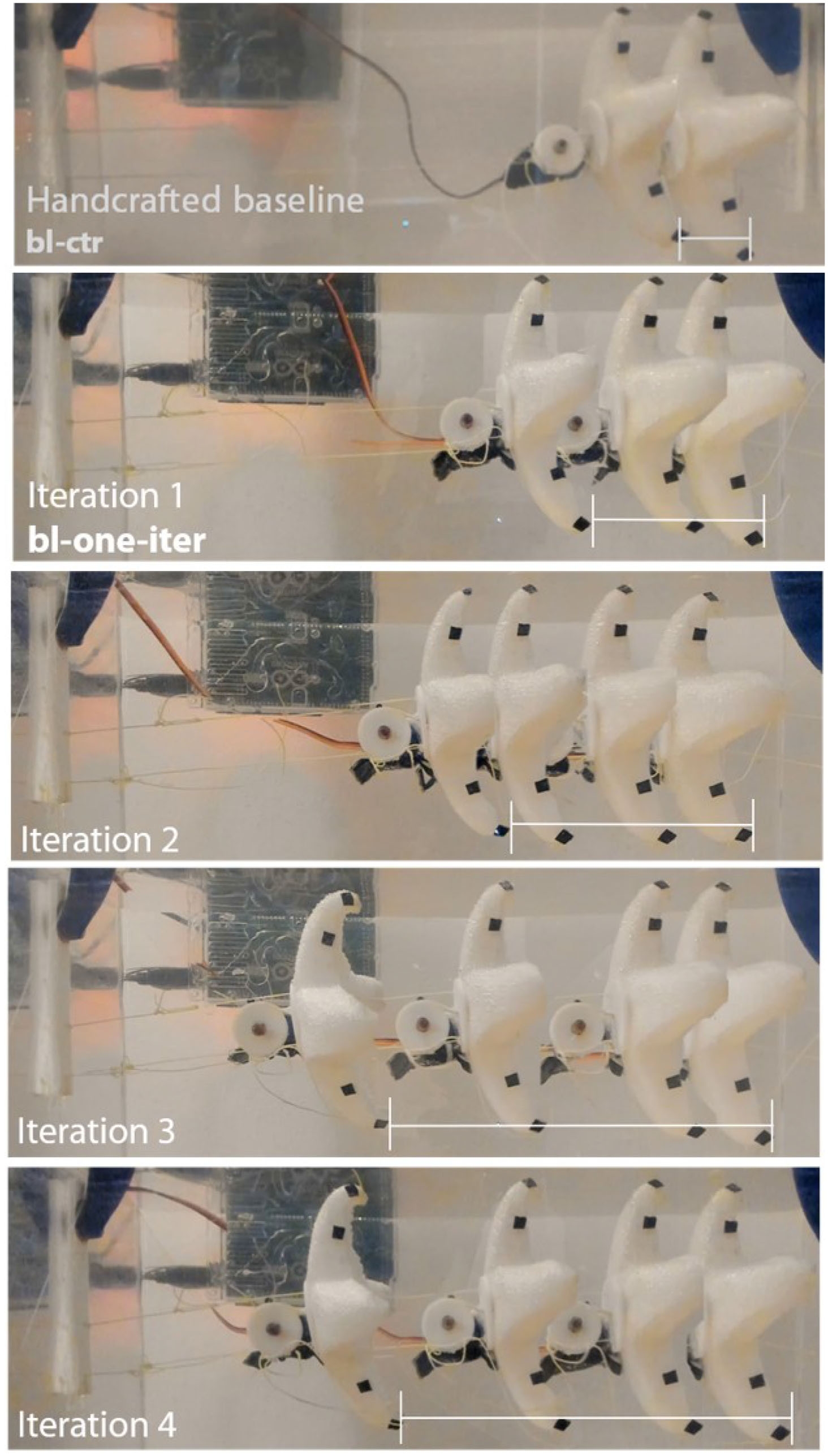}    
    \caption{(Left) \citet{turpin-2023-fast-simulation} optimize a grasp pose for various metrics. (Right) \citet{du-2021-underwater-simulation} use trajectory optimization to control the starfish to move forwards. The results are shown over several iterations. {\small(\textsuperscript{\textcopyright} 2024 IEEE)}}
    \label{fig:traj_opt}
\end{figure}

Trajectory optimization aims to determine optimal sequences of states and control inputs for dynamic systems. The goal is to minimize a predefined cost function, typically composed of a per-step loss function $\mathcal{L}(u_i, \hat{q}_i)$ and a terminal cost $\mathcal{L}_{\text{terminal}}(u_N, x_N)$. Here, $u \in \mathbb{R}^m$ represents control parameters, and $\hat{q}_1(u), \ldots, \hat{q}_N(u) \in \mathbb{R}^n$ denote the state trajectory estimated by the physics simulator given control parameters $u$ \ie,
\begin{subequations}
    \begin{align}
        \minimize_{u} \quad & \sum_{i=1}^{N-1}~\mathcal{L}(u_i, \hat{q}_i) + \mathcal{L}_{\text{terminal}}(u_N, q_N) \\
        \subjto \quad & u_{min} \leq u \leq u_{max},
        \end{align}
\end{subequations}
where $u_{min},u_{max}\in\mathbb{R}^m$ represent the minimum and maximum allowable values of the control parameters. This definition aligns closely with the system identification problem, however, focusing on optimizing the control signal under the assumption of known model parameters. Trajectory optimization has commonly been coupled with other tasks such as system identification~\citep{du-2021-underwater-simulation,lutter-2021-differentiable-learning,granados-2022-model-physics,chen-2022-real-simulation,wang-2022-real2sim2real-engine,wang-2021-sim2sim-robots} and morphology optimization~\citep{Xu-2021-An-Design, hu-2019-chainqueen-robotics}. For example, \citep{du-2021-underwater-simulation} combines system identification and trajectory optimization, by simultaneously updating both the trajectory and the model (against real-data) iteratively until a controller can produce large forward velocities (see \cref{fig:traj_opt}, right).

\textbf{Common Tasks:} A common trajectory optimization task involves moving a simulated body in a specified direction~\citep{hu-2019-chainqueen-robotics, wang-2022-real2sim2real-engine, wang-2022-recurrent-data, du-2021-diffpd-dynamics}, with the aim of either maximizing speed or maintaining a fixed speed. This task is applicable to various body types, such as soft bodies~\citep{hu-2019-chainqueen-robotics}, rigid-bodies~\citep{du-2021-diffpd-dynamics} and tensegerity robots~\citep{wang-2022-real2sim2real-engine, wang-2022-recurrent-data}. Besides moving in a fixed direction, studies have explored other simple motions like jumping~\citep{werling-2021-fast-contact} or lifting a leg on a quadruped robot~\cite{lelidec-2022-augmenting-smoothing}. \citet{wang-2023-softzoo-environments} presented benchmark tasks for the control of soft body objects, including moving forward, turning, tracking a velocity and following a set of waypoints.

Exploring the dynamics of throwing objects has been investigated in the context of robotic manipulators~\citep{geilinger-2020-add-contact, zhong-2021-extending-models}. A related task involves manipulating fabrics, where studies focus on placing a cloth into a bucket by applying forces to its corners~\citep{qiao-2020-scalable-control, gong-2022-fine-fabrics}. Additionally, \citet{li-2022-diffcloth-contact} investigated cloth-based tasks, specifically a dressing task using a manipulator to put socks and hats onto a simplified model of a human body.

Various manipulation-based tasks have also been explored, including reaching using muscle driven arms~\citep{wang-2022-differentiable-musculotendons}, trajectory optimization for specified paths~\citep{geilinger-2020-add-contact, petrik-2022-learning-physics}, object pushing~\cite{song-2020-learning-simulations} and batting a ball towards a target~\citep{werling-2021-fast-contact}. \citet{turpin-2022-grasp-hands} focuses on object grasping, optimizing hand poses through trajectory optimization. The approach in \cite{turpin-2022-grasp-hands} only requires the final pose after optimization, and this is extended in \cite{turpin-2023-fast-simulation} to enhance both speed and stability (see \cref{fig:traj_opt}, left). \citet{le-2023-differentiable-objects} explores robot trajectory optimization with contact interactions, specifically in a push-and-slide task where a simulated robot arm moves an object (Stanford bunny) to a goal while returning the robot's end effector to another goal, utilizing Dojo~\citep{howelll-2022-dojo-robotics}.

\citet{lin-2022-diffskill-tools} focus on tool-assisted manipulation tasks with soft bodies, like lifting and spreading dough on a cutting board, followed by flattening it with a rolling pin. In a similar context, \citet{li-2023-dexdeform-physics} use a robotic hand to manipulate dough towards specific goals. Generalizing the scope, \citet{huang-2021-plasticinelab-physics} present benchmarks for soft-body manipulation tasks involving agents reshaping plasticine material using manipulators. 


\textbf{Comparisons Against Other Methods:} \citet{huang-2021-plasticinelab-physics} demonstrate that gradient-based methods, by optimizing open-loop control sequences using differentiable physics engines, efficiently solve benchmark tasks, outperforming reinforcement learning (RL)-based approaches. Similar findings are reported in other works, including~\citep{hu-2019-chainqueen-robotics, liang-2019-differentiable-problems, qiao-2020-scalable-control, Xu-2021-An-Design, gong-2022-fine-fabrics}, highlighting the effectiveness of trajectory optimization compared to RL algorithms like proximal policy optimization (PPO)~\cite{schulman-proximal-2017-algorithms}. However, some tasks, such as those in \citet{li-2023-dexdeform-physics} and considered by \citet{lin-2022-diffskill-tools}, suggest that RL might outperform trajectory optimization, especially in scenarios with longer time horizons. Both studies recognize the value of differentiable simulation, whether in refining trajectories \citep{li-2023-dexdeform-physics} or aiding in data collection \citep{lin-2022-diffskill-tools, lv-2022-samrl-rendering}. Comparisons to derivative-free optimization techniques, commonly CMA-ES~\citep{hansen-2003-reducing-adaptation}, consistently show that gradient-based optimization achieves solutions faster and often produces superior results~\citep{geilinger-2020-add-contact, qiao-2020-scalable-control, werling-2021-fast-contact, Xu-2021-An-Design, wang-2022-differentiable-musculotendons}.

While trajectory optimization requires more computation time compared to RL, which can generate real-time reactive policies, a strategy to enhance its speed is to combine it with imitation learning over an optimized trajectory. \citet{chen-2022-imitation-physics} integrate a differentiable physics simulator into the policy learning computational graph, aiming to minimize the divergence between expert and agent trajectories. Consequently, this RL-based method aims emulate the characteristics of trajectory optimization while still offering real-time solution.


\subsection{Morphology Optimization}

Morphology optimization aims to find a set of material and geometric properties of a system that optimizes predefined objectives, subject to physical and geometric constraints. There are parallels between morphology optimization, system identification and trajectory optimization, since morphology optimization can either identify the optimal morphological parameters of the system with respect to a fixed trajectory~\citep{heiden-2019-interactive-simulation} or optimize them together with control parameters to generate a co-optimized trajectory~\citep{hu-2019-chainqueen-robotics, Xu-2021-An-Design, li-2023-learning-simulation}. The morphological parameters $\omega\in\mathbb{R}^p$, where $p$ is the number of morphological parameters, encompass physical characteristics such as shape, size, material properties, joint configurations, and other structural features that define the morphology of the agent under consideration. 


\textbf{Co-Optimizing Morphology and Control: }
A common application of morphology optimization is in co-design tasks, formulated as simultaneously optimizing the structural design of a robotic system and minimizing the cost of a control policy which drives the system towards a desired goal state. It is typical to minimize a per-step loss function $\mathcal{L}(u_i, \hat{q}_i, \omega)$ and a terminal cost $\mathcal{L}_{\text{terminal}}(u_N, q_N, \omega)$, over both the control parameters $u$ and the morphological parameters $\omega$, and the state trajectory estimated by the physics simulator $\hat{q}_1(u, \omega),\ldots,\hat{q}_N(u, \omega) \in \mathbb{R}^n$ given both the control and morphological parameters, \ie,
\begin{subequations}
    \begin{align}
        \minimize_{u, \omega} \quad & \sum_{i=1}^{N-1}~\mathcal{L}(u_i, \hat{q}_i, \omega) + \mathcal{L}_{\text{terminal}}(u_N, q_N, \omega) \\
        \subjto \quad & \omega_{min} \leq \omega \leq \omega_{max}, \quad u_{min} \leq u \leq u_{max},
        \end{align}
\end{subequations}
where $\omega_{min},\omega_{max}\in\mathbb{R}^p$ and $u_{min},u_{max}\in\mathbb{R}^m$ are the minimum and maximum allowable values of the morphological and control parameters respectively. Here, the loss function closely follows the definition of that for a typical trajectory optimization problem, however, the co-design task is defined over both the control parameters $u$ and the morphological parameters, $\omega$.

\begin{figure}
    \centering
    \begin{subfigure}{\linewidth}
	\includegraphics[width=\linewidth]{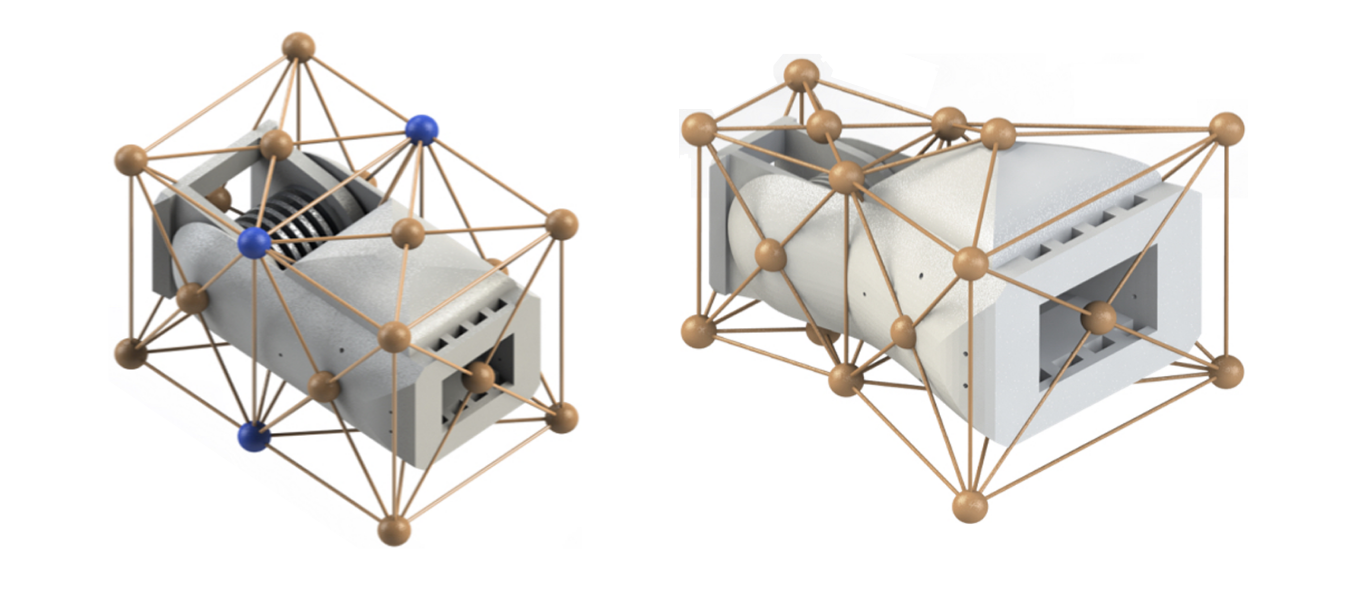}
        \caption{Visualization of the morphology optimization process of a robot joint and body segment from \cite{Xu-2021-An-Design}. (Left) Lower-dimensional morphology parametrization is constructed prior to optimization by posing deformation constraints on each component and joining their handle points (highlighted in blue) to ensure a feasible optimized morphology (Right).}
        \label{fig:cage-morph}
    \end{subfigure}\vspace{0.5cm}\\
    \begin{subfigure}{\linewidth}
	\includegraphics[width=\linewidth]{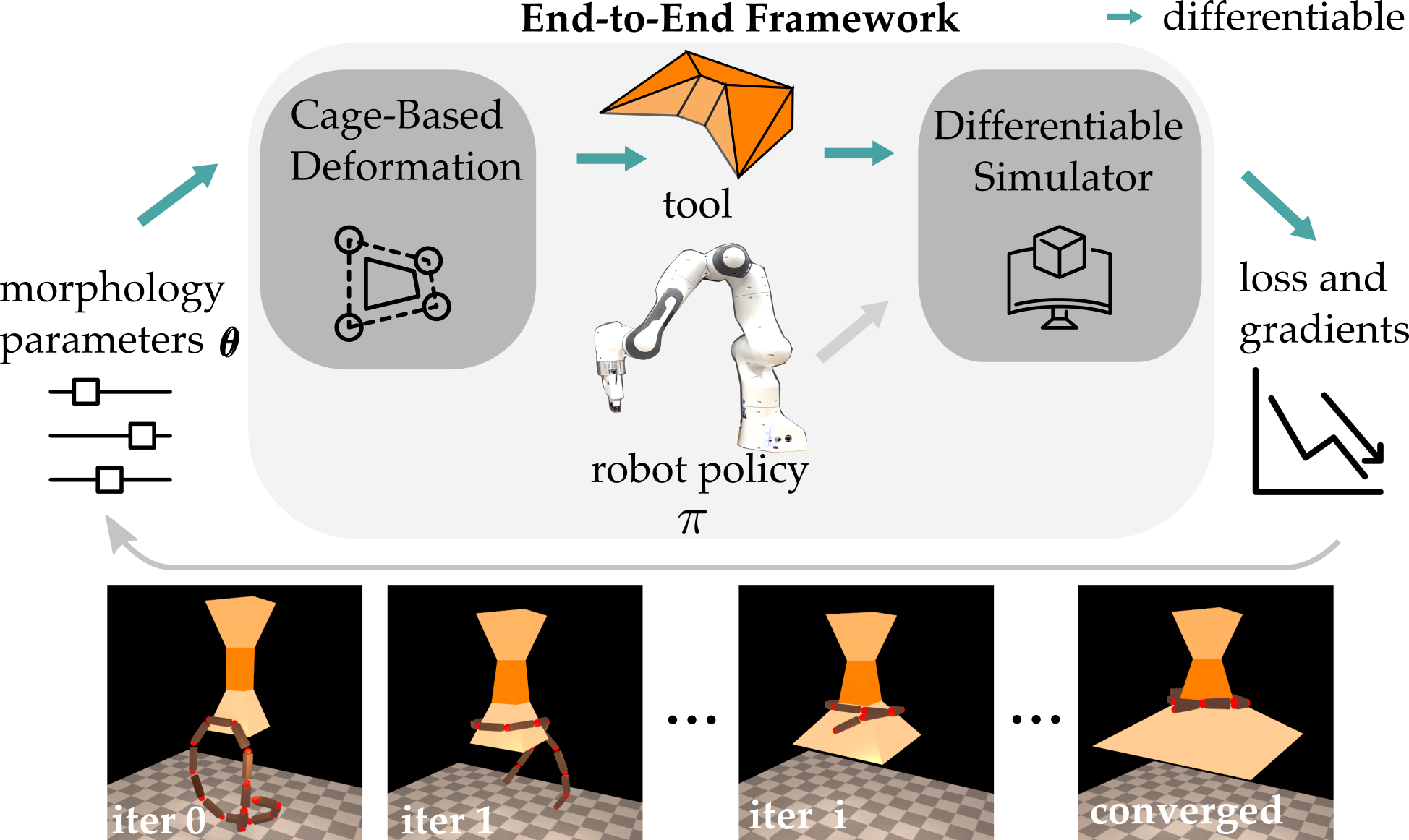}
        \caption{Overview of the end-to-end morphology optimization pipeline from \cite{li-2023-learning-simulation} leveraging differentiable simulation, where the tool morphology uses a similar caged-based parametrization as \cite{Xu-2021-An-Design}. At each iteration, the robot executes the task using the current tool parameters in simulation, after which gradients of the simulation are used to update the tool morphology parameters for the next iteration. {\small(\textsuperscript{\textcopyright} 2024 IEEE)}}
        \label{fig:tool-morph}
    \end{subfigure}
    \caption{Examples of differentiable simulation applied to morphology optimization. Both examples utilize caged-based morphology parametrization.}
    \label{fig:morph-opt}
\end{figure}

Differentiable simulation offers a means to overcome the \textit{curse of dimensionality} typically associated with such problems by allowing gradient information to be used when solving the problem. Because of the ability to make difficult problem spaces tractable, differential co-design \cite{choi2024dismech} is a popular methodology for optimizing morphology and control of soft robots.

Exemplar tasks include finding the optimal distribution of material stiffness (Young's modulus) across a deformable robot while minimizing the actuation energy to achieve a target pose \citep{hu-2019-chainqueen-robotics}, and optimizing the design of robot end-effectors and customized tools to enhance performance in contract-rich object manipulation tasks \citep{Xu-2021-An-Design, li-2023-learning-simulation} (see \cref{fig:morph-opt}). Other works have explored using a combination of gradient-based methods and graph search in the morphological space to optimize the design of autonomous underwater and aerial vehicles \citep{zhao-2022-graph-robots, zhao-2022-automatic-grammar} for performing highly-dynamic motions in simulated surveying tasks. \citet{heiden-2019-interactive-simulation} investigate the problem of optimizing the Denavit-Hartenberg parameters of a 4-DOF robot arm, given a pair of joint-space and task-space trajectories generated from an unknown kinematic mapping. Additionally, \citet{mezghanni-2022-physical-modeling} use the simulation gradients with respect to shape morphologies to train a neural network which generates more physically-stable shapes.


\textbf{Advantages and Challenges: }
Numerous works have found the use of differentiable simulators in co-design tasks to yield faster convergence and better constraint satisfaction \citep{hu-2019-chainqueen-robotics, Xu-2021-An-Design} compared to evolutionary algorithms and RL methods. \citet{li-2023-learning-simulation} show the robustness of the learned tool morphology on a sequence of manipulation tasks used in training with different initial conditions, however highlight the challenge of generalizing to other unseen tasks. Another drawback comes from the existence of local minima even in simple morphological design-spaces \citep{wang-2023-softzoo-environments}. 

\textbf{Improving the Optimization Process: } An existing method of enhancing the efficiency of gradient-based morphology optimization is to alter the parametrization model. Cage-based parametrization \citep{nieto2012cage} is a technique in computer graphics which uses vertices of a coarse, closed cage to control the enclosed space's deformation (see \cref{fig:morph-opt}). This method has been shown to result in much lower-dimensional optimization spaces \citep{Xu-2021-An-Design, li-2023-learning-simulation} compared to mesh-based counterparts, enabling gradient-based methods to more efficiently generate smoother morphology with comparable task performance than solutions obtained from mesh-based parametrizations. Through exploring the relationships between morphology representation, co-design tasks, and optimization algorithms, \citet{wang-2023-softzoo-environments} find that having a prior distribution in the design-space can further improve the effectiveness of the optimization.

\subsection{Policy Optimization} \label{subsec:policy-opt}

\begin{figure}[tb]
\centering
    \begin{subfigure}{\linewidth}
		\includegraphics[width=\linewidth]{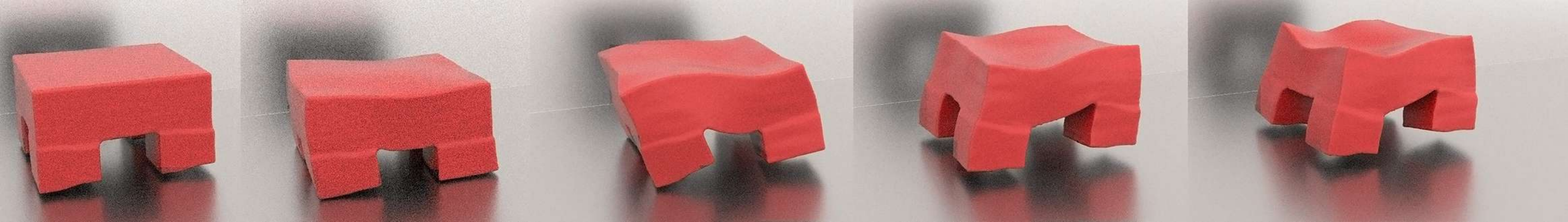}
        \caption{A soft-body 3D quadrupedal agent with 16 actuators, 4 per leg, learns a running policy using APG \cite{hu-2019-chainqueen-robotics}.}
        \label{fig:policy_opt1}
    \end{subfigure}
    \begin{subfigure}{\linewidth}
        \includegraphics[width=1.0\linewidth]{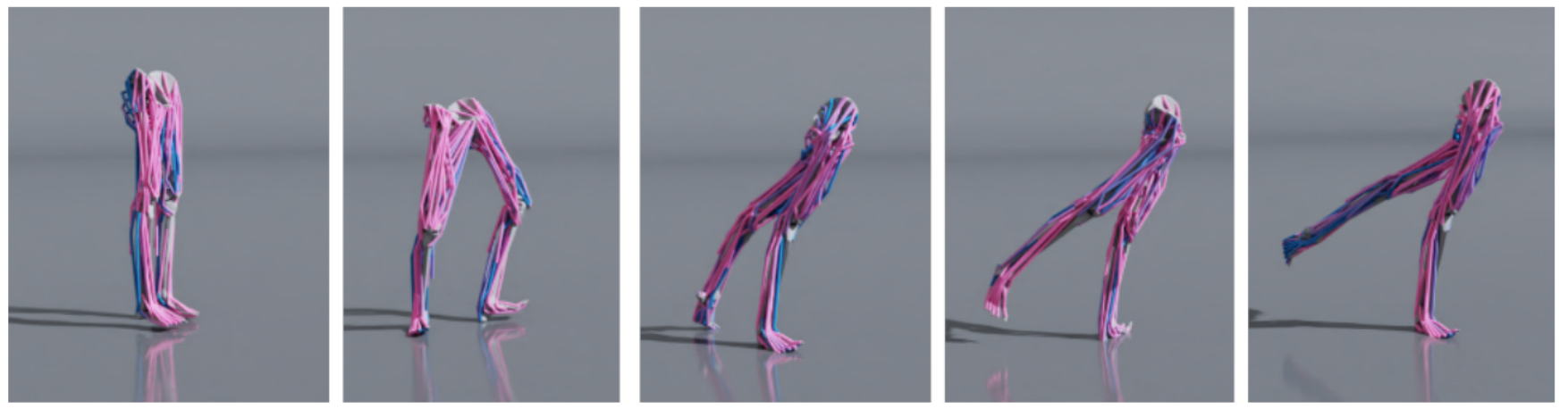} 
        \caption{Visualization of a policy learned using a differentiable simulator to control 152 muscle-tendon units to accomplish a running gait \cite{xu-2021-accelerated-simulation}.}
        \label{fig:policy_opt2}
    \end{subfigure}
    \caption{Examples of policies learnt using gradients available from differentiable physics simulators. In both cases the policies are parameterized indirectly with either a simple perceptron or a neural network.}
\end{figure}

In contrast to trajectory optimization (\cref{Trajectory Optimization}), a control policy abstractly represents actions conditioned on an environment's state \citep{sutton-2018-reinforcement-introduction}. However, policy optimization has received less attention in the context of differentiable simulation due to inherent local minima and discontinuities in the optimization landscape \citep{xu-2021-accelerated-simulation}. Direct parameterizations of control policies include optimizing the angular frequency parameter within a parametric curve~\cite{nava-2022-fast-models} and the parameters of a sinusoidal wave~\citep{heiden-2021-disect-cutting}. In the first example where the policy was conditioned on the angular frequency, the policy governed the motion of a soft carangiform swimmer while the parameterisation of the sinusoidal wave controlled the cutting action of a knife.  While successful for specific problems, representing control policies as parametric curves or sinusoidal waves is only feasible for select problems. For most tasks, neural networks offer greater flexibility but introduce complexity in the loss landscape. 


\textbf{Analytic Policy Gradients: } Some works optimize neural network parameters directly to encode control policies, termed Analytic Policy Gradients (APG) or policy optimization using Back Propagation Through Time (BPTT)~\cite{wiedemann-2023-training-gradient}. Examples of this are typically constrained to environments with limited to no variation in start state, open-loop control, simplified environmental contacts, and short horizons \citep{degrave-2019-a-robotics,geilinger-2020-add-contact,hu-2019-chainqueen-robotics,qiao-2020-scalable-control,spielberg-2023-advanced-chainqueen,ren-2023-diffmimic-physics,du-2021-diffpd-dynamics,li-2022-diffcloth-contact,gong-2022-fine-fabrics,xu2022efficient} (see Fig.~\ref{fig:policy_opt1} for an example). In closed-loop control tasks with long horizons or many contact-rich scenarios~\cite{lin-2022-diffskill-tools, antonova-2023-rethinking-perspective}, simple policy optimization methods like APG may yield suboptimal solutions. Therefore, various approaches are explored to find high-performing policies in such settings.


\textbf{Enhanced Policy Learning: } \citet{qiao-2021-efficient-bodies} proposed two techniques to enhance policy learning with differentiable physics engines: sample enhancement and policy enhancement. Sample enhancement leverages first-order approximation of gradients from a differentiable simulator to generate additional accurate samples, aiding the critic in learning a value function. However, with the advent of highly parallelized differentiable simulators \cite{macklin-2022-warp-graphics}, this approach may be less valuable than parallel data collection. In policy enhancement, the policy network is updated using physically-aware gradients from the differentiable physics simulator, which are generally more accurate. These methods were demonstrated on a pendulum control task and the classic ant control problem, showcasing significant improvements over state-of-the-art RL algorithms.

Policy Optimization via Differentiable Simulation (PODS) \cite{mora-2021-pods-simulation} extends Deterministic Policy Gradient (DPG) RL algorithms. It incorporates derivatives from a differentiable simulator by computing the analytic gradient of a policy's value function with respect to its actions. This allows for the refinement of trajectory actions, leading to improved estimates of the value function. The policy network can thus be updated using the difference in the initial and the improved trajectory. Both first- and second-order policy improvement are proposed, involving a line search for stability and, for the latter, computation of a Hessian, which can be computationally intensive. Reported results demonstrate faster convergence to high rewards compared to other state-of-the-art (SOTA) methods on pendulum and cable-driven manipulation control tasks.

Short Horizon Actor Critic (SHAC)~\cite{xu-2021-accelerated-simulation} is an RL algorithm leveraging gradients from a differentiable simulator over short time sequences, demonstrating performance improvements compared to SOTA RL algorithms (see Fig.~\ref{fig:policy_opt2}). The authors showcase SHAC's speed and sample efficiency on tasks including cartpole, ant, humanoid, and humanoid muscle-tendon unit (MTU), outperforming SOTA RL algorithms, first-order PODS, and BPTT in both wall time and sample efficiency. The method's success is credited to the GPU-based differentiable physics engine, enabling efficient simulation and accurate gradient updates to the actor within a truncated learning window, resulting in a smooth approximation in the loss landscape. However, when applied to the deformable manipulation tasks in DaxBench, SHAC was empirically worse than proximal policy optimization (PPO) and was often outperformed by APG. 

\citet{huang-2022-variational-physics} also proposes to do RL utilizing gradients from a differentiable simulator. However, uniquely they formalize the RL policy as a generative model over the distribution of optimal trajectories and train it using a variational lower bound optimized with gradients from both the classical policy gradient theorem and the differentiable simulator. The method's performance is only qualitatively evaluated on toy problems.

\textbf{Policy Learning using Imitation Learning: } In the domain of Imitation Learning, several works leverage differentiable physics simulation for trajectory optimization, with Behavior Cloning used to learn a policy parameterized as a neural network from expert trajectories \cite{collins-2021-follow-realitygrad, zhu-2023-difflfd-rendering,lv-2022-samrl-rendering}. In \citet{chen-2022-imitation-physics}, gradients are computed using an expert trajectory and a neural network policy where a Chamfer-$\alpha$ loss is proposed that calculates the distance between an expert trajectory and the rolled-out policy trajectory. 

\textbf{Improving Convergence: } Several methods have been proposed to enhance convergence in challenging optimization landscapes, yet their application to complex closed-loop control environments remains unexplored. Randomized smoothing \cite{lelidec-2022-augmenting-smoothing} addresses non-smoothness and non-convexity issues by using a noise distribution to evaluate gradients near the current optimization stage. Similarly, an $\alpha$-order gradient estimator \cite{suh-2022-do-gradients} interpolates between zeroth and first-order gradient estimates to tackle nonlinearities and discontinuities in differentiable physics, preventing convergence to suboptimal policies.


BO-Leap \cite{antonova-2023-rethinking-perspective} enhances a global search method with local search and gradient-based optimization, combining Bayesian Optimization, CMA-ES-like local search, and gradient-based optimization using differentiable simulator parameters. While effective in overcoming some discontinuities, it may struggle with non-smooth landscapes. Successful applications on tasks like cartpole and soft-body manipulation have been demonstrated, though the learned policies in these cases are open-loop with direct parameterizations.

\textbf{Summary: } In summary, most policy optimization methods attempt to learn neural network-parameterized control policies, with approaches like APG being common but prone to failure in increasingly complex tasks due to discontinuities and non-smoothness. While PODS and SHAC address some of these challenges, their robustness across diverse tasks remains unclear, leaving the integration of gradients from differentiable simulation for policy optimization an open question. Imitation learning approaches are often akin to trajectory optimization with behavior cloning. Promising yet under-explored directions involve incorporating randomized smoothing, adaptive interpolation of first and zeroth-order gradients, and augmentating gradient-based methods with global search algorithms to learn robust policies.







\subsection{Neural Network Augmented Simulation}

\begin{figure}
    \centering
    \includegraphics[width=\linewidth]{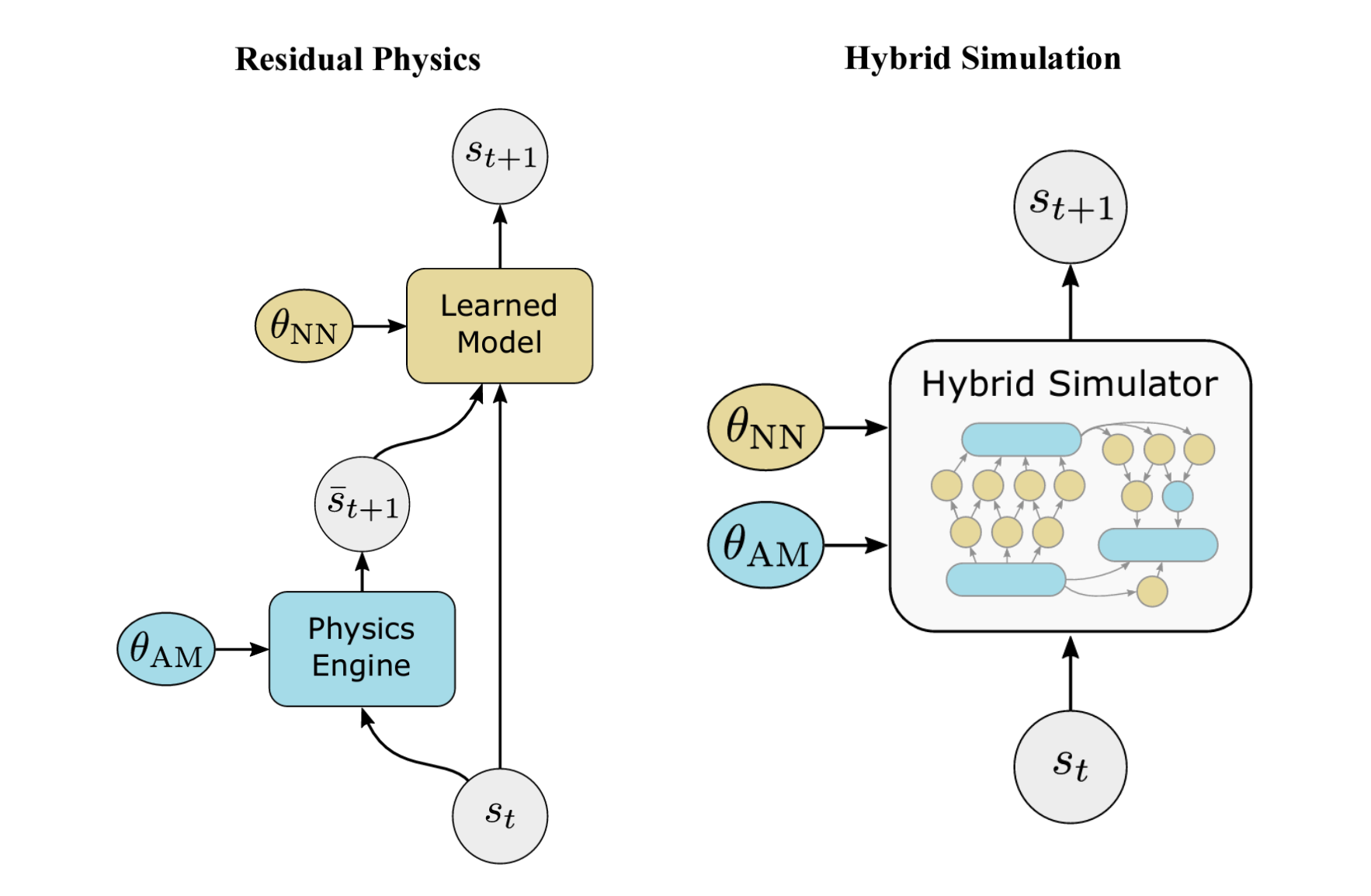}
    \caption{Comparison of neural network augmented simulation methods from \cite{heiden-2021-neural-networks}. (Left) The neural network learns the residual between the real states at the next timestep and simulated states from the physics engine. (Right) Hybrid simulation enables parts of the physics simulation to be replaced or augmented by neural networks, which can learn dynamics which the analytical physics models do not account for. {\small(\textsuperscript{\textcopyright} 2024 IEEE)}}
    \label{fig:hybridsim}
\end{figure}

Neural network augmented simulation involves the integration of a neural network and a differentiable physics engine for improved alignment between simulation and the real-world, while keeping the overall augmented simulation pipeline differentiable. \citet{lv-2022-sagci-learning} trained a neural network to learn the residual between the real future state and the simulated next state from the physics engine (\cref{fig:hybridsim}, left). Other works have explored embedding a differentiable physics engine within an autoencoder architecture, enabling end-to-end training of the overall network. This has been applied to system identification \citep{heiden-2019-interactive-simulation, de-2018-end-control}, where object states decoded from input RGB frames are propagated forwards by a small timestep through the physics engine, before being encoded to generate output frames. These works have demonstrated the advantage of neural network augmented simulation resulting in better alignment between real and simulated environments. 

Others used a neural network to learn a model of key physical parameters such as frictional contact dynamics \citep{heiden-2021-neural-networks} and physical properties of objects \citep{de-2018-end-control} (\cref{fig:hybridsim}, right). These works have shown the augmented simulation approach to yield faster convergence on estimated system parameters that are closer to ground-truth, enabling better generalization and more accurate long-term rollout predictions than deep-learning approaches, while accounting for dynamics that the physics engine does not model \citep{heiden-2021-neural-networks}.


\section{Discussion}
\label{sec:discussion}

The increasing interest in differentiable simulators primarily arises from the current demands of machine learning paradigms, which heavily rely on gradients for optimization. The proliferation of efficient gradient computation techniques and an improved understanding of gradient-based optimization algorithms have further fueled this trend. As detailed in Section~\ref{sec:differentiable_simulators}, there exists a strong array of differentiable simulators, serving as a valuable resource for newcomers interested in exploring the capabilities of these physics simulators or for integration into existing workflows. 

The applications of differentiable simulators, as explored in Section~\ref{sec:applications}, encompass a diverse range of areas including the optimization of simulation parameters such as actions, dynamics, and morphologies, as well as integration into broader differentiable workflows. However, the journey toward fully realizing the potential of these simulators is marked by notable challenges. \citet{suh-2022-do-gradients} illuminate several critical limitations of current differentiable simulators. In complex physical systems, particularly those involving long-horizon planning and control, the performance of differentiable simulators is hindered by landscape characteristics such as non-linearity, non-smoothness, and discontinuities. This complexity underscores the necessity of nuanced approaches in gradient estimation and optimization.

Moreover, using deterministic gradients from these simulators can lead to sub-optimal behavior due to the inherent landscape characteristics. The phenomenon of `empirical bias', where approximations of discontinuous dynamics lead to inaccuracies in gradient estimates, presents another significant challenge. Additionally, high variance in gradient estimates can be problematic, especially in scenarios with persistent stiffness or chaotic dynamics.

These complexities not only highlight the necessity for thoughtful design and algorithmic considerations in developing differentiable simulators but also underscore the importance of ongoing research and refinement in this evolving field.

\subsection{Future Directions}

Looking ahead, the path for differentiable simulators includes several promising areas of research and development. Some of these areas are application driven whilst others call for further research into the accuracy of simulation whilst enhancing the availability of meaningful gradients. 



We see the following research directions as being areas of growth and continued development for differentiable simulators:

\subsubsection{Better Gradient Estimation}


Contact models (\cref{subsec:contact}) are a key component of differentiable simulators, however, existing implementations only rely on simplified approximations of real contact dynamics through compliant models or optimization-based models for ease of implementation and differentiation. It is currently not clear how these simplifications affect the differentiable simulation performance, both in terms of the quality of generated gradients and how well the simulated behavior translates to the real world. This motivates further research to investigate improving the gradients available through proposing better contact models \cite{zhong-2023-improving-contacts}. Potential alternative approaches to obtaining more accurate or useful gradients include methods such as interpolating between zeroth and first-order gradients \cite{suh-2022-do-gradients}, or using a perturbed~\cite{werling-2021-fast-contact} or averaged gradient over a probabilistic distribution about a point~\cite{lelidec-2022-augmenting-smoothing,zhu-2023-difflfd-rendering}.

\subsubsection{Leveraging Gradient Information}


Effectively leveraging the available gradients is another area of exploration. Due to the limitations of current differentiable simulators and the gradients they provide, methods that augment local gradient-based optimizations with global search strategies such as Bayesian optimization or evolutionary algorithms is a promising path towards finding high performing solutions of nonconvex optimization problems. An example of this is \cite{antonova-2023-rethinking-perspective} where three levels of optimization are used to find high performing parameterized policies. Similarly, \cite{gartner-2022-differentiable-reconstruction} utilizes basin-hopping, which combines a local search (\eg, gradient descent) with random perturbations to escape local minima.
Policy optimization as an application area is comparatively under-researched and due to the slower than real-time speed of trajectory optimization using a differentiable simulator \cite{zhu-2023-difflfd-rendering}, policy optimization will remain an important area of research in the future. Being able to utilize the gradients available from the simulator to find high-performing policies parameterized using a neural network, similar to \cite{xu-2021-accelerated-simulation} will become increasingly important in future as it will allow for online, reactive policies to be developed. Finally, exploring the use of different loss functions or novel parametrisation methods \cite{li-2023-learning-simulation} to better leverage gradient information and enhance computational efficiency are also promising directions worth investigating. 



\subsubsection{Long Horizon Tasks}

Optimization over long horizons is an open problem across research fields with different communities experiencing unique aspects of this problem and proposing solutions accordingly, \eg reinforcement learning \cite{Nachum2018Data-EfficientLearning}, motion planning in robotics \cite{yamada2023leveraging} and recurrent neural networks \cite{Dasgupta2013OnNetworks}. Differentiable simulators share some common challenges associated with long time horizons, \eg vanishing and exploding gradients \cite{xu-2021-accelerated-simulation}, but also have some unique problems. For example, the large increase in memory and computational cost when calculating gradients over extended simulations. To assist the associated memory cost, checkpointing \cite{qiao-2021-efficient-bodies} has been proposed. Also, as the gradients from a differentiable simulator only provide local information, this often causes the optimization of long-horizon trajectories to find local optima \cite{yamada2024dcubed}. DiffSkill \cite{lin-2022-diffskill-tools} overcomes local optima by breaking the long-horizon problem down by using intermediary goals and skill abstraction. Although, some aspects of long-horizon optimization problems are shared between fields, we argue that some aspects of this problem will be unique to users of differentiable simulators.  Consequently, we foresee this research direction gaining prominence in the future, particularly as the complexity and duration of simulated tasks continue to increase.




\subsubsection{Differentiable Sensor Simulation}
The advancement of differentiable simulators can be enhanced by the incorporation of differentiable sensor models like those offered in \citet{sundaresan-2022-diffcloud-objects}, \citet{xu2022efficient} and \citet{jatavallabhula-2021-gradsim-control}. These models facilitate a robust real-to-sim and sim-to-real workflow, crucial for tasks like environmental perception and interaction in robotics. Incorporating approaches such as differentiable rendering of images and point clouds, through frameworks like PyTorch3D \cite{ravi-2020-accelerating-pytorch3d}, not only augments the realism and practicality of simulations but also preserves their differentiable nature. This trend paves the way for further exploration into diverse sensor models and the integration of multiple differentiable sensors into simulators, mirroring real-world sensor capabilities. 

\subsubsection{Online Applications}

Optimizing actions online in real-time using gradients from differentiable simulators is an appealing concept. However, the present-day differentiable simulators operate at or below real-time speeds \cite{zhu-2023-difflfd-rendering}, limiting their application within online optimization frameworks. This research direction, explored by \citet{chen-2022-real-simulation}, holds immense potential, though the real-world tasks demonstrated so far have been constrained to relatively simple operations due to the current limitations of simulators.


\subsubsection{Real-to-Sim Transfer Learning}

The domain of real-to-sim transfer learning, where comprehensive environmental models are reconstructed within simulators to facilitate downstream tasks like planning and trajectory optimization, is another application for differentiable simulators we expect to see grow in the future \cite{sundaresan-2022-diffcloud-objects, le-2023-differentiable-objects}. This direction, encompassing a spectrum of modeling challenges beyond those addressed in system identification (\cref{sec:sysID}), has witnessed notable advancements in representing environmental dynamics within simulators while leveraging their differentiable properties \cite{zhu-2023-difflfd-rendering, lv-2022-sagci-learning}.

\subsubsection{Challenges for Widespread Adaptation}

Differentiable simulation offers several advantages over current simulation suites. Open-source simulators like Brax \cite{freeman-2021-brax-body}, Warp \cite{macklin-2022-warp-graphics}, and TDS \cite{heiden-2019-interactive-simulation} support parallel simulation, leveraging GPU capabilities for simultaneous environments, which has been utilized in the literature for rapid model training in tasks like imitation learning \cite{chen-2022-imitation-physics}. Furthermore, most differenitable simulators seamlessly integrate into common workflows such as PyTorch or JAX, facilitating their incorporation into specific environments. However, challenges persist, including limited sensing capabilities, perceived complexity with a steeper learning curve for APIs, and comparatively fewer features than established simulation software. Moreover, the necessity of gradients is not universally established, especially in reinforcement learning applications, necessitating further research. Additionally, optimization techniques tailored to the nuanced landscapes of differentiable simulation require development to fully realize its potential.


\section{Conclusion}

This review has highlighted the significant advancements and diverse applications of differentiable simulators, underscoring their transformative role in computational physics and machine learning. With the aim of answering the questions: what are differentiable simulators, how do they work and how have they been used in research thus far, this review has explored the field, covering the foundations and the core components of differentiable simulators. Looking to the future of differentiable simulators, it is evident that the continued development and refinement of differentiable simulators will be pivotal in advancing fields that rely on physically informed gradients such as robotics, control, machine learning and others.


\bibliographystyle{myieee}
\bibliography{references}
\begin{IEEEbiography}[{\includegraphics[width=1in,height=1.25in,clip,keepaspectratio]{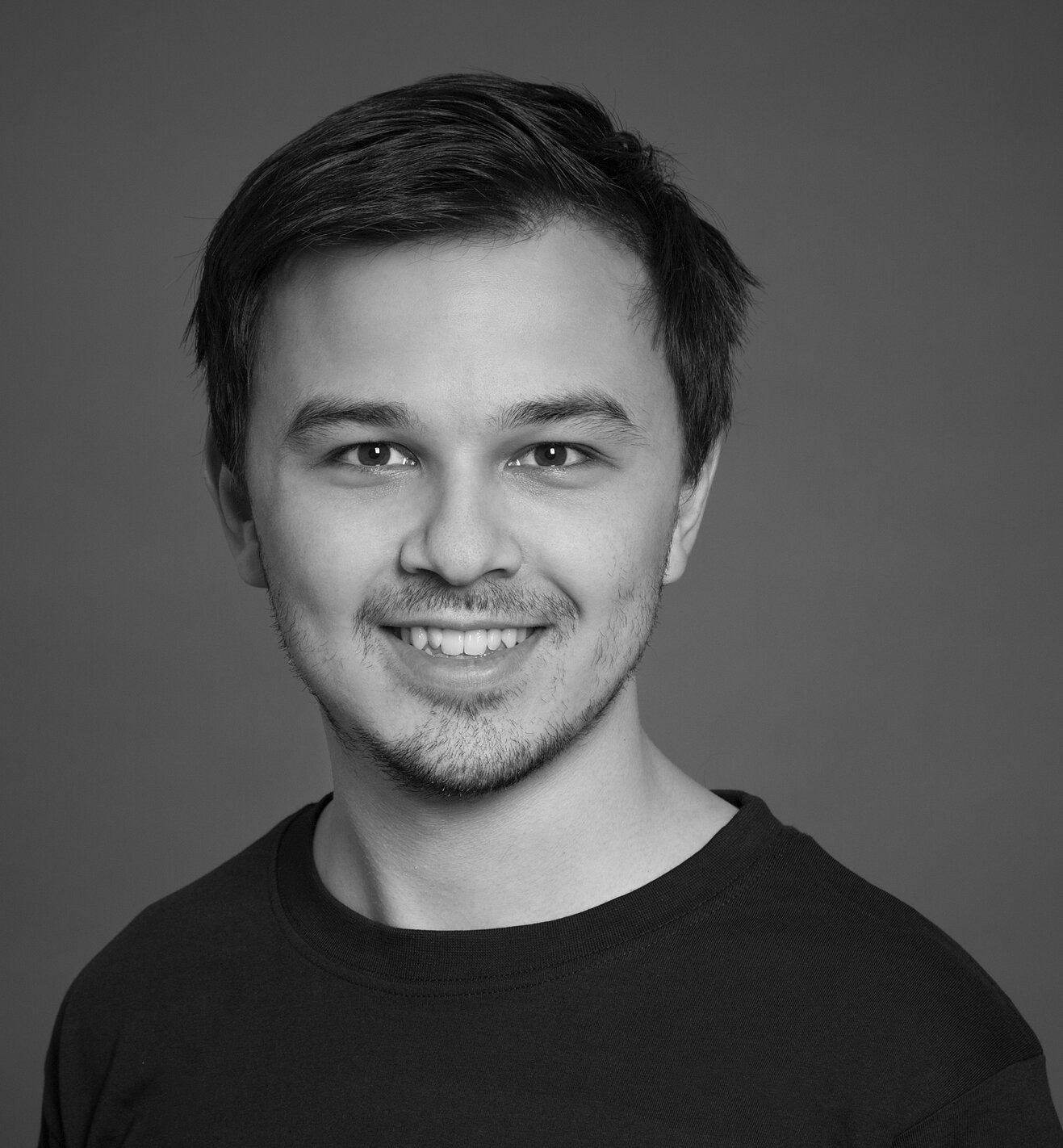}}]{Rhys Newbury} is a PhD student at Monash University, Australia. He holds a B. Eng (Hons.) in mechatronic engineering and B. Sci (Computer Science) from Monash University. His research interests focus around manipulation and using reinforcement learning.
\end{IEEEbiography}

\begin{IEEEbiography}[{\includegraphics[width=1in,height=1.25in,clip,keepaspectratio]{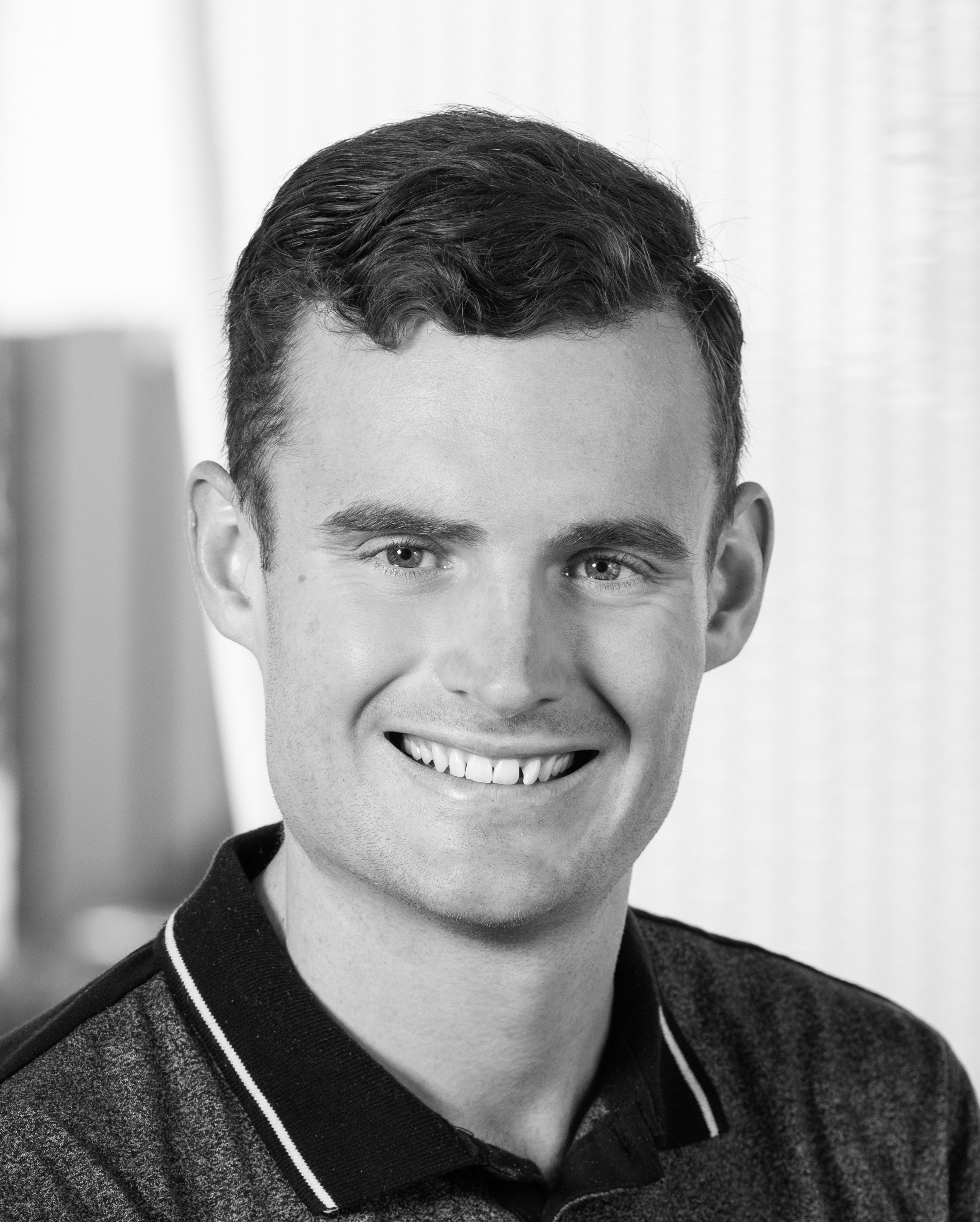}}]{Jack Collins} received the B.Eng. (Hons.) degree in mechatronic engineering from the Queensland University of Technology, Australia, in 2017 and the Ph.D degree in robotics from the Queensland University of Technology, Australia, in 2022. Since 2021 Jack has been a Postdoctoral Researcher at the Oxford Robotics Institute in the Applied AI Lab, University of Oxford. His research interests include simulation and the sim-to-real gap, representation learning for scene understanding and prediction, learning from demonstration and task and motion planning for long-horizon tasks.
\end{IEEEbiography}

\begin{IEEEbiography}[{\includegraphics[width=1in,height=1.25in,clip,keepaspectratio]{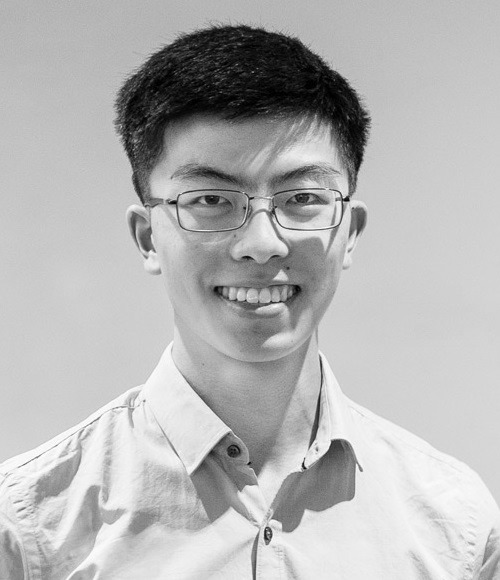}}]{Kerry He} is a PhD candidate with the Department of Electrical and Computer System Engineering at Monash University. His current research is about convex optimization methods for problems arising from quantum information theory. He received his BEng (Honors) and BComm from Monash University, during which he worked on optimal control and trajectory planning for robotic manipulators and driverless vehicles.
\end{IEEEbiography}

\begin{IEEEbiography}[{\includegraphics[width=1in,height=1.25in,clip,keepaspectratio]{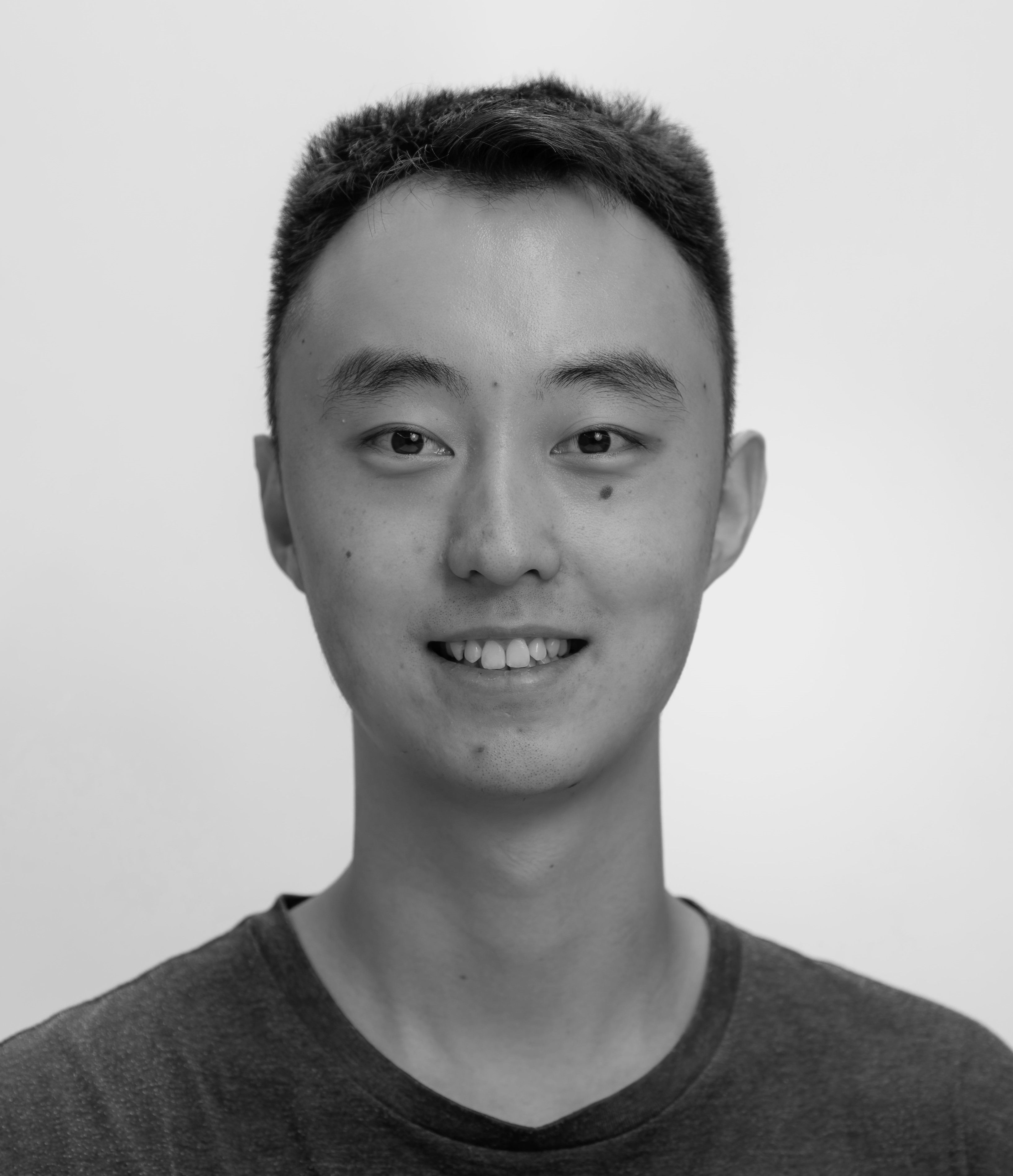}}]{Jiahe Pan} is a research assistant in the Human-Robot Interaction group at The University of Melbourne, Australia. He received the B.Sc. degree in Mechatronics Engineering from The University of Melbourne, in 2023. His current research is in designing adaptive autonomous robots for human-robot collaboration.
\end{IEEEbiography}

\begin{IEEEbiography}[{\includegraphics[width=1in,height=1.25in,clip,keepaspectratio]{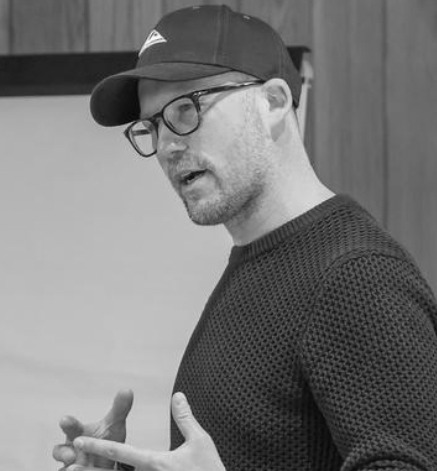}}]{Ingmar Posner} leads the Applied Artificial Intelligence Lab at Oxford University and is a founding director of the Oxford Robotics Institute. His research aims to enable machines to robustly act and interact in the real world - for, with, and alongside humans. With a significant track-record of contributions in machine perception and decision-making, Ingmar and his team are thinking about the next generation of robots that are flexible enough in their scene understanding, physical interaction and skill acquisition to learn and carry out new tasks. His research is guided by a vision to create machines which constantly improve through experience. In 2014 Ingmar co-founded Oxa, a multi-award winning provider of mobile autonomy software solutions. He currently serves as an Amazon Scholar.

\end{IEEEbiography}

\begin{IEEEbiography}[{\includegraphics[width=1in,height=1.25in,clip,keepaspectratio]{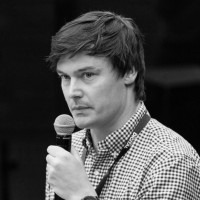}}]{David Howard} received the B.S. and M.Sc. degrees from the University of Leeds, U.K., in 2005 and 2006, respectively, and the Ph.D. degree from the University of the West of England, U.K., in 2011. He has been with the Commonwealth Scientific and Industrial Research Organization, Brisbane, Australia, since 2013. His research interests include embodied cognition, the reality gap, and soft robotics.
\end{IEEEbiography}

\begin{IEEEbiography}[{\includegraphics[width=1in,height=1.25in,clip,keepaspectratio]{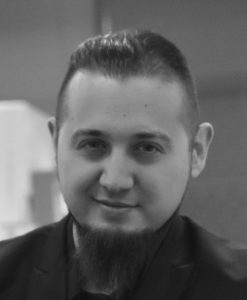}}]{Akansel Cosgun} is a Senior Lecturer at Deakin University, Australia. He received his Ph.D. degree in robotics from the Georgia Institute of Technology, USA, in 2016. From 2018 to 2022, he was a Research Fellow at Monash University, Australia. He conducts research in robotics, human–robot interaction, and robot learning. He has previously worked with Honda Research, Toyota Infotechnology Center, Microsoft Research, and Savioke. His research interests include mobile robots, robotic arms, and self-driving cars with an emphasis on a systems view to problems.
\end{IEEEbiography}

\EOD
\end{document}